%% file: root.tex
\documentclass[letterpaper, 10 pt, conference]{format/ieeeconf}  

\IEEEoverridecommandlockouts                              

\usepackage{graphicx}
\usepackage{subcaption} 
\usepackage{epsfig} 
\usepackage{amsmath} 
\usepackage{amssymb}  
\usepackage{tikz}
\usepackage{siunitx}
\usepackage{multirow}
\usepackage{pifont}
\usepackage{url}
\usepackage{xcolor} 
\usepackage{hyperref}
\usepackage{cleveref}
\usepackage{tabularray}

\usepackage[citestyle=numeric-comp, sorting=none, maxbibnames=99]{biblatex}
\AtBeginBibliography{\small}

\definecolor{mygreen}{RGB}{0,128,0}
\definecolor{myred}{RGB}{198,0,0}
\usetikzlibrary{backgrounds}
\newcommand{\cmark}{\ding{51}}%
\newcommand{\xmark}{\ding{55}}%

\usepackage[normalem]{ulem}
\useunder{\uline}{\ul}{}

\tikzset{%
  every neuron/.style={
    circle,
    draw,
    minimum size=0.3cm,
    color=brown,
    fill=brown!50
  },
  neuron missing/.style={
    draw=none, 
    scale=1,
    text height=0.333cm,
    execute at begin node=\color{black}$\vdots$,
    fill=none
  },
}

\usetikzlibrary{math}

\title{\Large \bf
Vision Transformers for End-to-End Vision-Based Quadrotor Obstacle Avoidance
}

\author{Anish Bhattacharya$^{*}$, Nishanth Rao$^{*}$, Dhruv Parikh$^{*}$, Pratik Kunapuli, Yuwei Wu, Yuezhan Tao,\\Nikolai Matni, Vijay Kumar
\thanks{*Equal contribution.}
\thanks{Authors are with The General Robotics, Automation, Sensing \& Perception (GRASP) Lab, University of Pennsylvania,
        3330 Walnut St, Philadelphia, PA 19104, United States. Corresponding author: {\tt\small anish1@seas.upenn.edu}.}%
\thanks{This work was funded by NSF SLES-2331880 and NSF CAREER-2045834.}}

\begin{document}    

\maketitle
\thispagestyle{empty}
\pagestyle{empty}

\begin{abstract}

We demonstrate the capabilities of an attention-based end-to-end approach for high-speed vision-based quadrotor obstacle avoidance in dense, cluttered environments, with comparison to various state-of-the-art learning architectures. Quadrotor unmanned aerial vehicles (UAVs) have tremendous maneuverability when flown fast; however, as flight speed increases, traditional model-based approaches to navigation via independent perception, mapping, planning, and control modules breaks down due to increased sensor noise, compounding errors, and increased processing latency. Thus, learning-based, end-to-end vision-to-control networks have shown to have great potential for online control of these fast robots through cluttered environments. We train and compare convolutional, U-Net, and recurrent architectures against vision transformer (ViT) models for depth image-to-control in high-fidelity simulation, observing that ViT models are more effective than others as quadrotor speeds increase and in generalization to unseen environments, while the addition of recurrence further improves performance while reducing quadrotor energy cost across all tested flight speeds. We assess performance at speeds of up to 7m/s in simulation and hardware. To the best of our knowledge, this is the first work to utilize vision transformers for end-to-end vision-based quadrotor control.

\end{abstract}


\setlength{\textfloatsep}{5pt plus 1.0pt minus 2.0pt} 
\setlength{\floatsep}{5pt plus 1.0pt minus 2.0pt} 
\setlength{\intextsep}{5pt plus 1.0pt minus 2.0pt} 
\setlength{\abovecaptionskip}{2pt} 
\setlength{\belowcaptionskip}{0pt} 

\section{INTRODUCTION}
\label{sec:introduction}

\input{introduction}

\section{RELATED WORK}
\label{sec:related-work}

\input{related-work}

\section{METHODOLOGY}
\label{sec:methodology}

\input{methodology}

\section{RESULTS}
\label{sec:results}

\input{results}

\section{CONCLUSION}
\label{sec:conclusion}

\input{conclusion}

\printbibliography 

\end{document}


\title{\textbf{Supplementary}\\\Large{Vision Transformers for End-to-End Vision-Based Quadrotor Obstacle Avoidance}}
\maketitle

\section{Code, datasets, and pretrained weights}

We make available all simulation and training code and datasets, pretrained weights for each model, and real robot experiment code (ROS1) at this GitHub: \url{https://github.com/anish-bhattacharya/ViT-for-quadrotor-obstacle-avoidance}.

\section{Results: Generalization to flight through window}

\begin{figure}[h]
    \centering
    \begin{subfigure}{.63\linewidth}
        \includegraphics[width=1.0\linewidth]{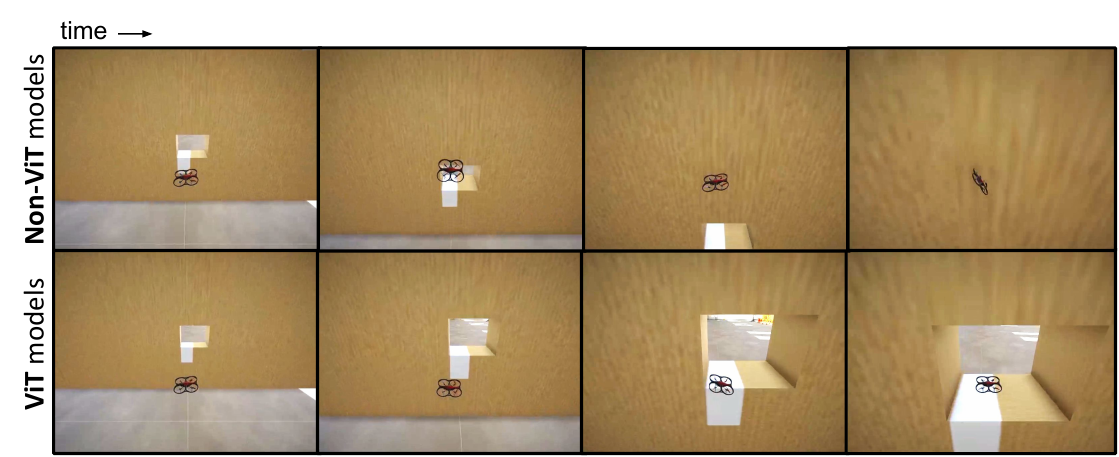}
        \caption{Representative simulation trials.}
        \label{fig:window-experiment-sim}
    \end{subfigure}
    \hfill
    \begin{subfigure}{0.348\linewidth}
        \includegraphics[width=1.0\linewidth]{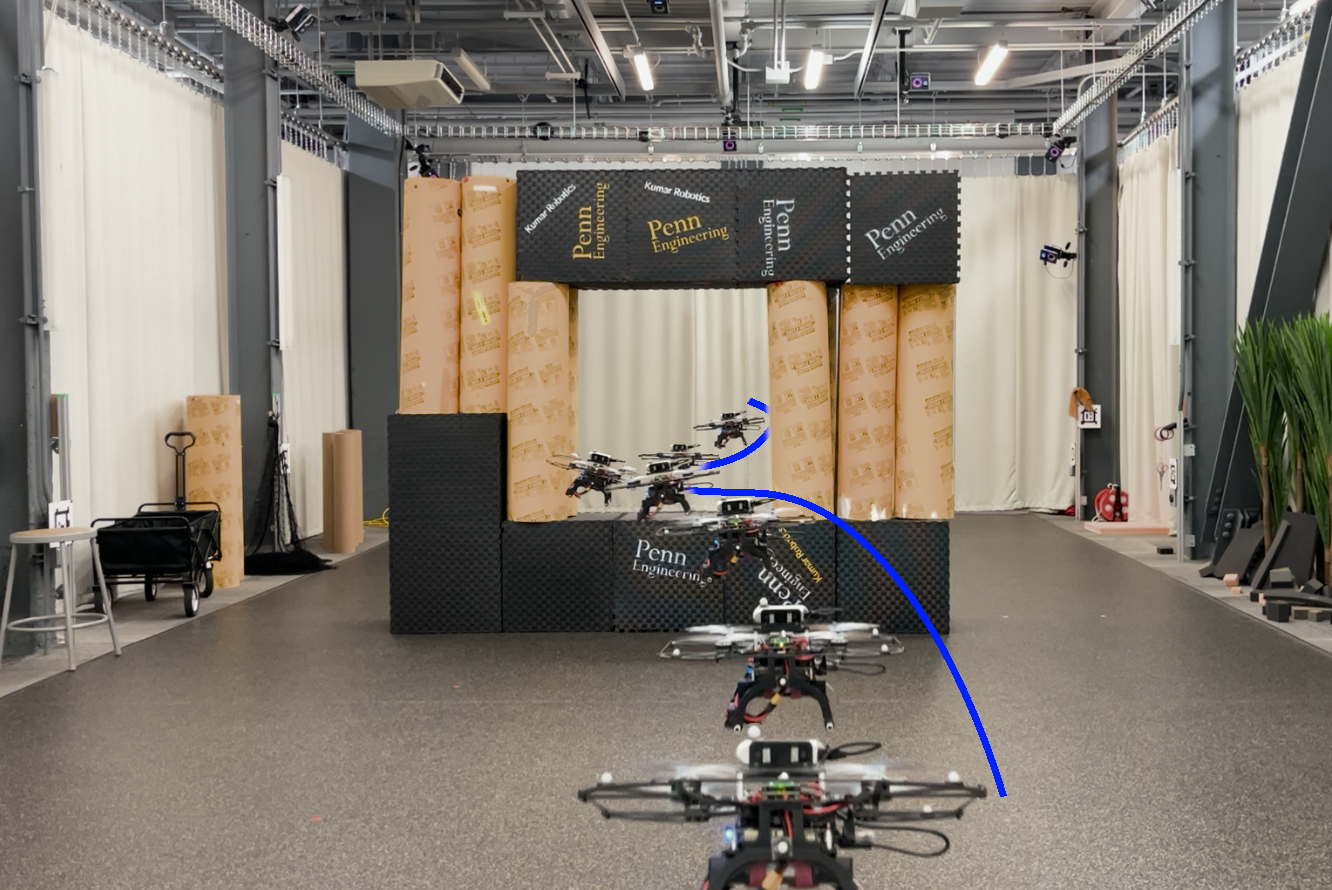}
        \caption{Real trial.}
        \label{fig:window-experiment-real}
    \end{subfigure}
    \caption{Demonstration of paths taken by non-ViT and ViT models for the \textit{simulated} flight-through-window task, as well as a \textit{real hardware} demonstration of this task. Only the ViT-based models succeed at this challenging task.}
    \label{fig:window-experiment}
\end{figure}

We additionally present a zero-shot fly-through-window experiment performed in simulation across all models at 3 m/s (Figure \ref{fig:window-experiment-sim}). While the Spheres and Trees environment contains obstacles within free space, with multiple viable collision-free paths, this environment has only a tight distribution of viable collision-free paths given a drone start position. Only the ViT models (ViT and ViT+LSTM) succeed at this task, and all others fail. This presents additional evidence that ViT models are superior in generalization compared to other state-of-the-art vision models, including recurrent models, for end-to-end quadrotor control.


We perform a real-world trial of the fly-through-window experiment, presented in Figure \ref{fig:window-experiment-real}, at 2m/s with the ViT+LSTM model and no additional fine-tuning.



\section{Results: Further quantitative metrics}

\input{supp-more-results}

\section{Results: Real attention maps for ConvNet vs. ViT+LSTM}

\begin{figure}[h]
    \centering
    \includegraphics[width=\linewidth]{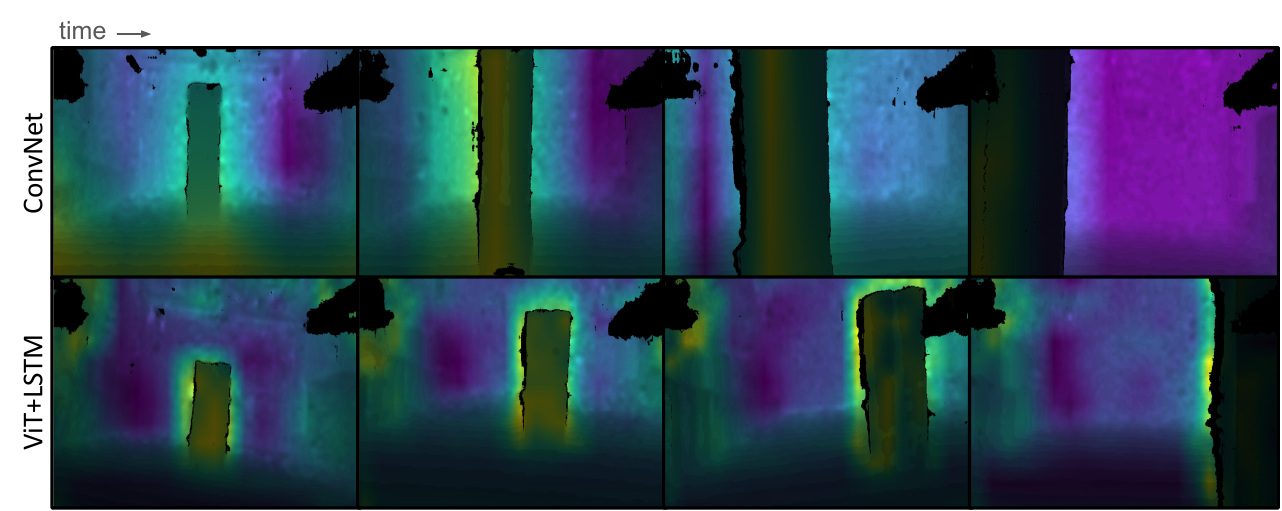}
    \caption{Attention maps generated from a real experiment using the ConvNet and ViT+LSTM model, respectively. Notably, the ViT+LSTM consistently highlights the relevant parts of the object (edges) throughout the flight. Triangular black regions in the top left and right of each image are propeller guards of the quadrotor on which depth is not accurately computed.}
    \label{fig:real-attmaps}
\end{figure}

Figure \ref{fig:real-attmaps} shows attention maps from the ConvNet and ViT+LSTM models, respectively, during the avoidance of a real-world, indoor obstacle. The ConvNet model produces a diffuse attention (spatial activation) over the obstacle which is not consistent over time, whereas the ViT+LSTM consistently highlights edges of the object during the approach and avoidance maneuver.

\section{Model network architectures}

We provide additional details about the five architectures presented in the main text of this paper. Table \ref{tab:model-layers} includes information about the specific structures of each model, and Figures \ref{fig:tikz-convnet}, \ref{fig:tikz-lstmnet}, \ref{fig:tikz-unetlstm}, \ref{fig:tikz-vit}, \ref{fig:tikz-vitlstm} contain corresponding diagrams. As described in the main text, each architecture except for ConvNet was designed to be close to 3M parameters, dictating, for example, the number of LSTM layers used for recurrent architectures.

\input{tables/table-model-details}

\include{tikz-nn-diagrams}




%% file: introduction.tex


Quadrotor unmanned aerial vehicles (UAVs) are small, agile vehicles that can fly with high levels of acceleration and agility. With the miniaturization of onboard sensing and computing, it is now common to treat quadrotors as a single, independent robot that senses its environment, performs high-level planning, calculates low-level control commands, then tracks these commands using an onboard-calculated state estimate. Prior to recent work, most research in this field has been done with the traditional robotics pipeline of individual and modular sensing, planning, and control blocks, each with some computational requirements and error tolerances. 
However, quadrotors reach their highest utility when flown fast, thereby achieving more coverage with limited battery life; 
at these speeds, perception-to-planning via the modular approach can break down \cite{falanga2019fast}. Recent works have explored using learned policies to develop a single sensing-to-planning algorithm that feeds into a tracking controller; this has fast reaction times and generalizes to sensor noise and real-world artifacts when learned models are trained with domain randomization techniques \cite{Loquercio_2021}. In contrast, we explore a fully end-to-end approach, encapsulating more of the tracking task into a vision transformer (ViT) model.




\begin{figure}[t]
  \centering
  \begin{subfigure}[t]{.45\linewidth}
    \centering
    \includegraphics[height=2.81cm]{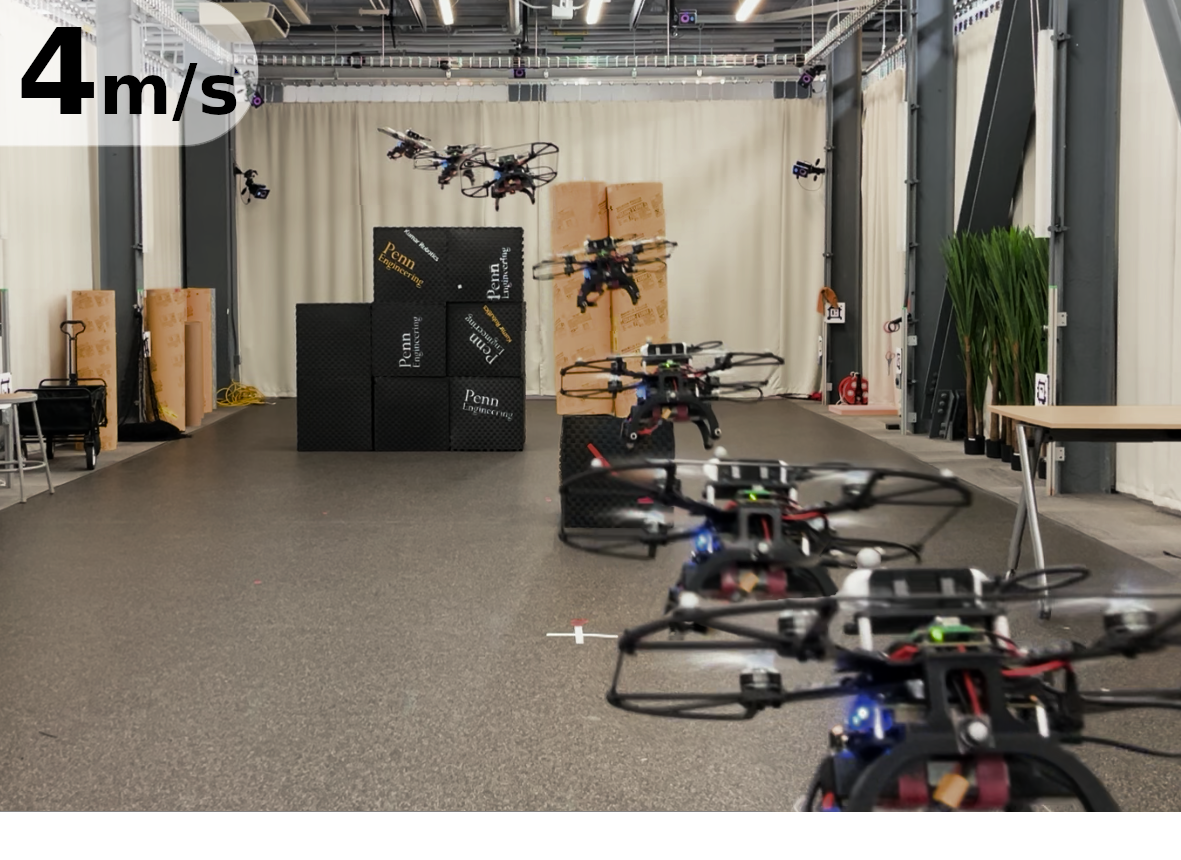}
    \vspace{-19pt}\caption{Evasive maneuver.}
    \label{fig:frontpage-3d-dodging}
  \end{subfigure}%
  ~
  \begin{subfigure}[t]{.53\linewidth}
    \centering
    \includegraphics[height=2.82cm]{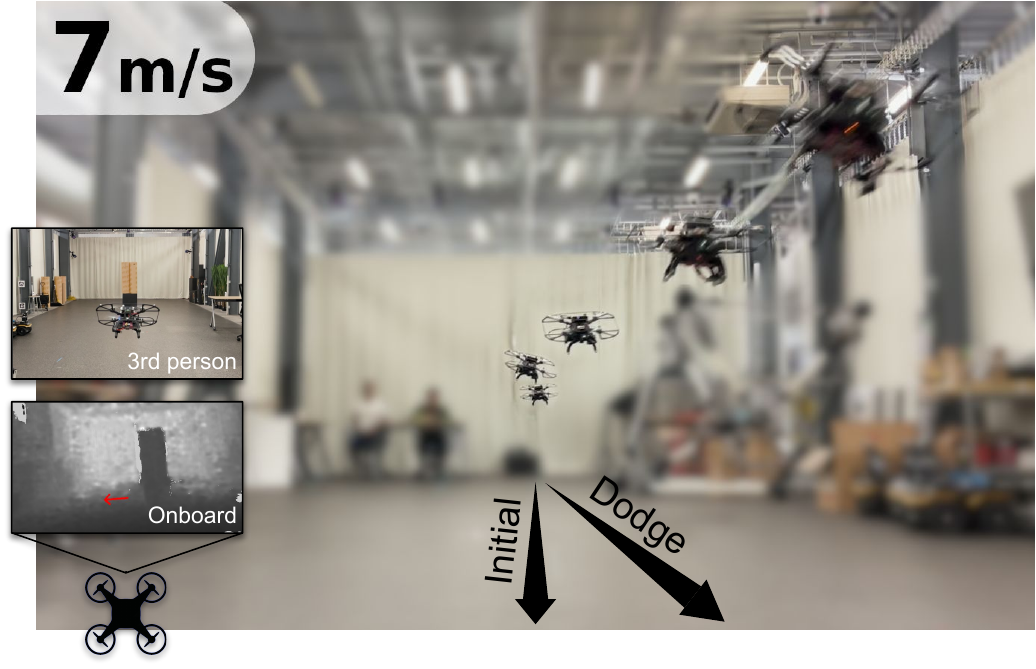}
    \vspace{-7pt}\caption{Fast-flying, oncoming drone.}
    \label{fig:frontpage-perspective-view}
  \end{subfigure}
  \vfill
  \begin{subfigure}[t]{1.0\linewidth}
    \centering
    \includegraphics[width=1.0\linewidth]{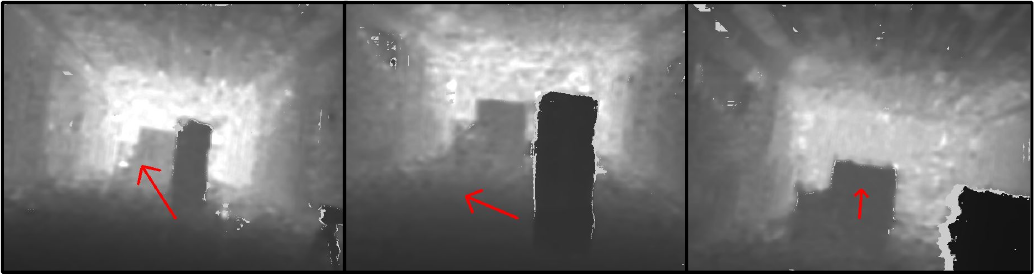}
    \vspace{-15pt}\caption{Onboard depth and velocity commands (red) sequence for \ref*{fig:frontpage-3d-dodging}.}
    \label{fig:frontpage-dbg-im}
  \end{subfigure}
  \caption{Zero-shot sim-to-real transfer for high-speed and multi-obstacle, 3D evasive maneuver with a combination vision transformer-recurrent model.}
  \label{fig:frontpage}
\end{figure}


We aim to explore the utility of attention-based learning models, specifically vision transformer architectures, that take depth images as input for end-to-end control of a quadrotor for the high-speed obstacle avoidance task. We adopt the task and simulator from ICRA 2022 DodgeDrone Challenge: Vision-based Agile Drone Flight \cite{dodgedrone-competition}, showcasing various teams' efforts to develop obstacle avoidance policies for a quadrotor flying through a cluttered simulation environment. 
Against the ViT architecture, we select as baselines a range of learning backbones representing varying use-cases: convolutional, U-Net, and recurrent networks. While this is not an exhaustive list of all popular learning architectures available today, we believe this covers a wide range of model types typically used for vision-based control-related tasks, including object detection, image segmentation, and temporal sequence modeling.

The favorable qualities of convolutional networks \cite{convnet-original-lecun, convnet-housedigits} have driven much of the work in vision-based UAV control. These networks are able to learn feature maps that, when trained with gradients computed relative to a desired objective function, outperform hand-engineered feature maps and additionally are translationally invariant. However, when used for the control of dynamical systems, they do not maintain any internal state and thus do not have any memory of past states or actions. In robotics tasks where both the robot and thus the sensory input are dynamic, it has yet to be shown that convolutional networks are the best method for perception. 
Recurrent architectures (e.g., long short term memory \cite{lstm}) are models that maintain an internal state and may provide smoother, more feasible commands to dynamic robots such as quadrotors \cite{ou2021autonomous}. More recent sequence modeling techniques use the attention mechanism \cite{attention} to construct transformers \cite{transformers} that have been adapted from the natural language processing domain for vision \cite{vits} and show superior performance in object recognition and tracking. These vision transformers take as input the patches of an image as a sequence, and when provided with large training datasets, outperform convolutional or recurrent networks in a growing number of computer vision tasks. Further motivation lies in prior work which suggests that using recurrent or attention-based models might improve performance due to the temporal nature of dynamic robot navigation \cite{song2023learning}.



This work's aim is to study the ViT as an end-to-end framework for quadrotor control, with comparisons to baseline architectures. We choose to train models via behavior cloning in simulation from a privileged obstacle-aware expert, as is common among existing learning-to-avoid works \cite{loquercio2018dronet, Loquercio_2021}. This additionally eliminates potentially confounding factors that may exist in a reinforcement learning framework, such as the exploration/exploitation tradeoff and additional hyperparameter tuning.
We use depth sensing data to train and evaluate these models, as across multiple sensor types, robotic platforms, and environments, depth data has shown in prior work to be both ubiquitous and effective at high-level scene understanding as well as navigation tasks. 
We evaluate models against each other by comparing collision metrics during forward flight through a cluttered scene, across trials with increasing forward velocity. We further present the paths taken by each model through a given simulated scene, the command characteristics of each model (acceleration and energy cost), and compare our end-to-end approach to modular baseline methods. Representative hardware experiments further demonstrate agile avoidance around multiple obstacles and at high speeds (Figure \ref{fig:frontpage}).




\noindent\textbf{Our contributions are as follows:}
\begin{itemize}
    \itemsep0em
    \item The first use of vision transformer models for high-speed end-to-end control of a quadrotor.
    \item A comparison of a vision transformer model against various state-of-the-art, learning-based architectures for end-to-end depth-based quadrotor control.
    \item Real experiments demonstrating and comparing models on hardware.
    \item Open-source code, datasets, and pretrained weights to reproduce and extend the results in this paper.\footnote{\url{www.anishbhattacharya.com/research/vitfly}}
\end{itemize}

%% file: related-work.tex

\noindent\textbf{End-to-end vision-based navigation.}
End-to-end approaches for control tasks have been investigated in a variety of settings such as autonomous driving \cite{pmlr-v87-mueller18a}, and partially so in autonomous quadrotor flight in the wild \cite{Loquercio_2021}. Both works demonstrate that such approaches have high sample complexity, and in the more structured self-driving task, it does not generalize as well as modular approaches.
However, both end-to-end and structured methods degrade in performance as agility increases. Existing works for learned end-to-end quadrotor flight largely focus on flight at a relatively slow speed and use raw images directly \cite{dai2020automatic, loquercio2018dronet}.


\noindent\textbf{Vision transformers for control.} Some research has used a combination of architectures to surpass the performance of individual components for the object detection task. Swin Transformer \cite{liu2021swin} and a ResNet-based module \cite{he2016resnet} have been combined previously \cite{hendria2023combining}, showing that this attention+convolutional network architecture improved object detection performance. We similarly explore combination models between basic backbone architectures (convolution and U-Net, recurrence, and attention) in this work. When considering downstream tasks such as control from pixels, transformer architectures such as ViT have been compared to convolutional neural networks (CNNs) and shown to perform worse than the CNN architecture due to the weaker inductive bias and need for more training data \cite{tao2022evaluating, li2023a}. ViT has been studied extensively for its use in downstream control tasks as a representation learner in many reinforcement learning (RL) works \cite{hansen2021stabilizing, kalantari2022improving, seo2023masked}. Our method investigates leveraging similar architectures using imitation learning from a privileged expert as opposed to RL. 


\noindent\textbf{Vision transformers with quadrotors.} Since the development of the ViT, some works have used such models with onboard imagery from quadrotors but not in the same module used for planning or control. Some perform object detection onboard a UAV \cite{reedha2022transformer, ye2023real}. Object tracking and then servoing has been attempted in the past, where a Siamese Transformer Network performs object tracking onboard a UAV, then independently a PD visual servoing algorithm commands yaw angle and forward velocity from the bounding box \cite{sun2022siamese}. To our knowledge, our work represents the first investigation for using ViT in an end-to-end manner with quadrotor control. 

\noindent\textbf{Quadrotor flight from depth images.} When navigating in uncertain environments, depth images offer an input modality for learned, reactive control \cite{Loquercio_2021, zhou2019robust, florence2020integrated} and can even be treated as an intermediate, learnable task for downstream control \cite{bhattacharya2024monocular}. Various approaches have been taken to learn control outputs from depth image inputs such as mapping then planning \cite{zhou2019robust}, using motion primitives \cite{florence2020integrated}, and sampling collision-free trajectories \cite{Loquercio_2021}. These methods use depth images to determine some intermediate representation which is then used to control the quadrotor. Compared to these methods, we demonstrate purely end-to-end control using various architectures learning from depth images directly, without the need for mapping, planning, or trajectory generation.

%% file: methodology.tex
\begin{figure}
    \centering
    \includegraphics[width=1.0\linewidth]{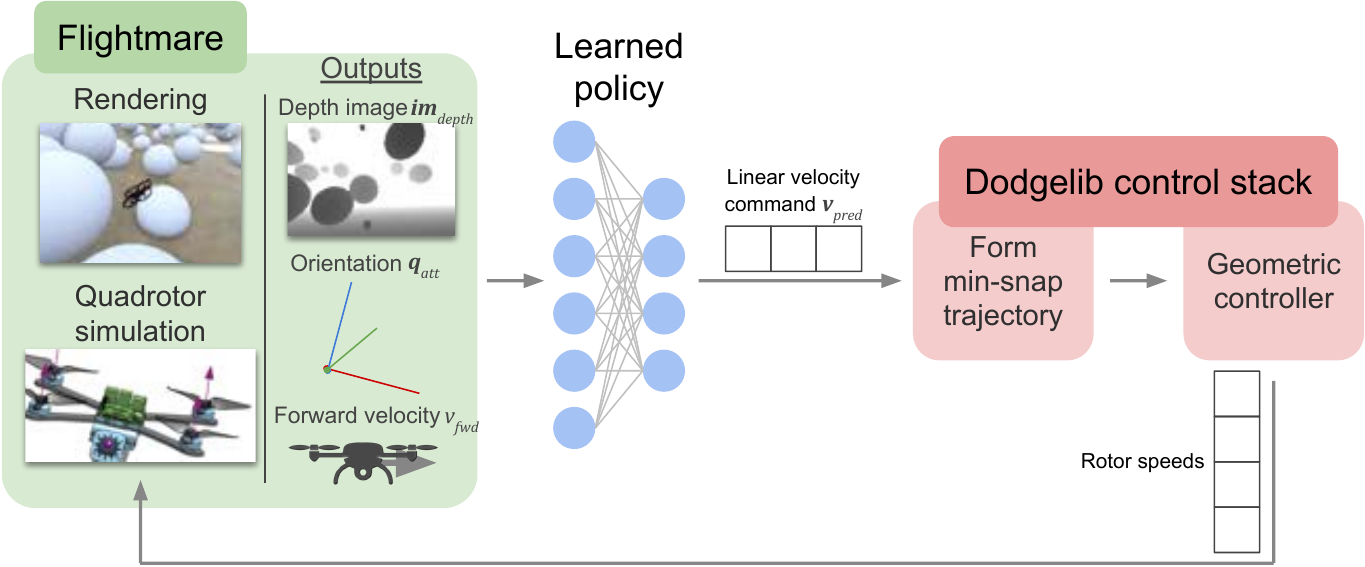}
    \caption{Depth images $\mathbf{im}_{depth}$ and quadrotor orientation $\mathbf{q}_{att}$ come from Flightmare, and along with forward velocity $v_{fwd}$ serve as input to the chosen learning model. The model outputs a linear velocity command $\mathbf{v}_{pred}$ which leads to the formation of a min-snap trajectory tracked by a geometric controller, both part of the Dodgelib control stack \cite{dodgedrone-competition}. This outputs rotor speeds, which is sent to the quadrotor simulator in Flightmare.}
    \label{fig:system-diagram}
\end{figure}

\subsection{Task Formulation}
\label{subsec:task-formulation}

We formulate the obstacle avoidance task as flying forward a fixed distance at various forward velocities while avoiding static obstacles in a cluttered environment that is unstructured and unknown. 
The privileged teacher expert policy is a reactive planner designed to be short-sighted and mimic data available to low-skill pilots, thereby avoiding expensive real-world, high-risk, in-situ data collection which may rely on trained-pilot data but be more time-optimal. Supervisory data is collected during expert policy rollouts to train the end-to-end learning-based student models. As this work aims to study ViTs against other models for developing reactive obstacle avoidance behavior, all models output linear velocity commands in the world frame, in contrast to long-horizon trajectory planning often used for less reactive policies in prior work \cite{Loquercio_2021}. To compare model performance, we consider multiple obstacle collision metrics, computational complexity and inference time, and other quantitative and qualitative factors (see Section \ref{sec:results}).

Formally, we consider an obstacle-aware expert $\pi_{\text{expert}}(s_t, o) = a_t$ taking as input quadrotor state $s_t$ and local obstacle locations $o$, and producing action $a_t$ at every timestep $t$. A dataset $\mathcal{D}$ is collected by conducting many trajectories $\tau$ under which the transition dynamics $\mathcal{T}$ are produced under the expert $s_{t+1} = \mathcal{T}(s_t, a_t), a_t \sim \pi_{\text{expert}}(s_t, o)$ and the depth image $\mathbf{im}_{\text{depth}, t}$ is recorded at every timestep. We seek to train a student policy $\pi_{\text{student}}(s_t, \mathbf{im}_{\text{depth}, t}) = a_t$ not directly aware of obstacle locations but rather relying on depth images to reproduce the actions of the expert policy \eqref{eq:loss}.

\subsection{Learning framework}
\label{subsec:learning-framework}

The student model takes as input a single depth image 
$\mathbf{im}_{depth} \in [0, 1]^{60 \times 90}$ (similar to \cite{Loquercio_2021}) of size 60$\times$90, quadrotor attitude $\mathbf{q}_{att}=(w, x, y, z)$, and forward velocity $v_{fwd}\in\mathbb{R}$, and predicts a velocity vector $\mathbf{v}_{pred}\in\mathbb{R}^3$ (see Figure \ref{fig:system-diagram}). As models have access to the forward velocity, we do not include variations of model input such as stacked image sequences; while this input might improve temporal predictions of convolution-only models, we instead present recurrent models that are meant to model sequential data. The $\mathbf{v}_{pred}$ are learned from privileged expert $\mathbf{v}_{cmd}$ via supervised behavior cloning, with a L2 loss over all timesteps in a trajectory, given student model parameters $\mathbf{\theta}$.

\begin{equation}
    L(\mathbf{\theta}) = \mathop{\mathbb{E}}_{\tau \sim \mathcal{D}}  \left[\frac{1}{T} \sum^{T-1}_{t=0} \left\| \mathbf{v}_{cmd}(t) - \mathbf{v}_{pred}(t;\mathbf{\theta}) \right\|^2_2 \right]
    \label{eq:loss}
\end{equation}

\subsection{Simulated setup and models}
\label{subsec:experimental-setup}

\noindent\textbf{Simulation and rendering.}
We use the open-source Flightmare simulator \cite{flightmare} that contains fast physics simulations of quadrotor dynamics and a bridge to the high-visual-fidelity game engine renderer Unity. In addition to being ROS-enabled, this simulator package allows us to collect training data from and deploy quadrotors in various environments with accurate depth sensor measurements with real-time computations on a laptop computer. Specifically, we use a modified version of the simulator supplied for the DodgeDrone ICRA 2022 Competition \cite{dodgedrone-competition} which contains an environment with floating spherical obstacles of various sizes (later referred to as ``Spheres" environment). The simulated quadrotor has a mass of 0.752kg and a diameter of 0.25m.

\noindent\textbf{Privileged expert policy and data collection.}
To gather supervisory data pairs of $\{(\mathbf{im}_{depth}, \mathbf{q}_{att}, v_{fwd}), \mathbf{v}_{pred}\}$ for behavior cloning, the privileged expert runs a short-horizon collision avoidance algorithm through randomized obstacle fields. Note that bold variables are multidimensional.
In total, 588 expert runs yielded 112k depth images, where 27.5k images are sampled from trials where there is at least one collision. Since the expert is reactive and lacks dynamics-feasible planning, collisions are not absent from the data; however, we find that some models (see Section \ref{subsec:metrics}) outperform this expert at high speeds.

\begin{figure}
  \centering
  \begin{subfigure}[t]{.45\linewidth}
    \centering
    \includegraphics[width=1.0\linewidth]{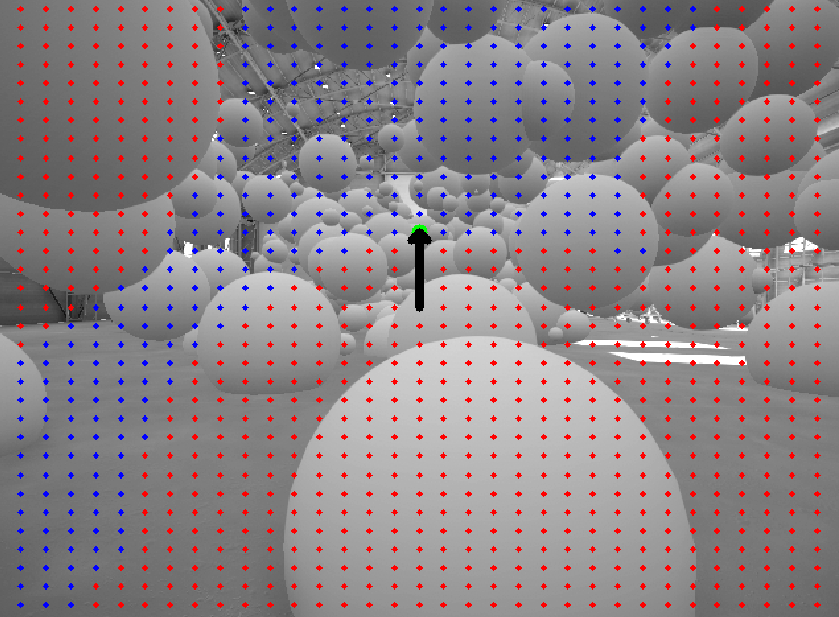}
    \vspace{-15pt}
    \caption{Expert waypoint selection.}
    \label{fig:expert-wpts}
  \end{subfigure}%
  ~
  \begin{subfigure}[t]{.45\linewidth}
    \centering
    \includegraphics[width=1.0\linewidth]{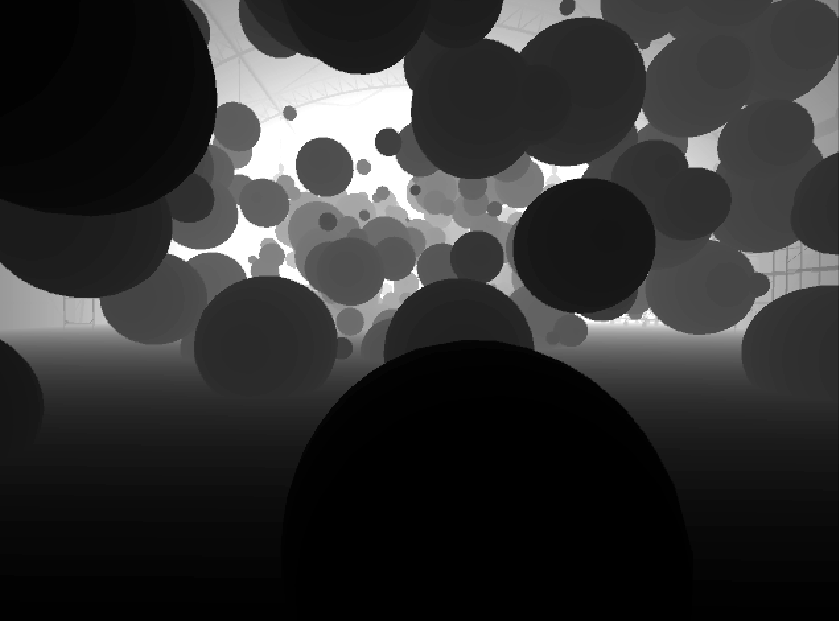}
    \vspace{-15pt}
    \caption{Collected depth image.}
    \label{fig:expert-depth}
  \end{subfigure}
  \caption{Expert policy visualized from the quadrotor onboard camera in a sample training environment. \ref*{fig:expert-wpts} shows waypoints collision-queried at a fixed horizon given privileged obstacle location information (red: in collision, blue: free) where the green represents the chosen waypoint for issuing a velocity command towards (indicated with an arrow). This velocity command and the corresponding depth image (\ref*{fig:expert-depth}), as well as quadrotor attitude and velocity, are the collected data at this timestamp.}
  \label{fig:expert-policy-vis}
\end{figure}

The privileged expert can access obstacle position and radius information $o$ of those obstacles within 10m of the current drone position $s_{t, pos}$. It searches for straight-line collision-free trajectories from the drone's current position to each waypoint along a 2D grid in the lateral-vertical plane set at some horizon ahead of the drone, as demonstrated in Figure \ref{fig:expert-policy-vis}. The waypoint closest to the center of the grid is chosen, and a gain is applied to this relative position to generate a velocity command action $a_t$. Transition dynamics $\mathcal{T}$ are provided by the Flightmare quadrotor simulation.

\noindent\textbf{End-to-end student models.}
We present five student models for a comparison of ViT-based models against other state-of-the-art image processing architectures. Each model takes as input a single depth image, and uses an encoder-like module composed of either convolutions or attention to generate a lower-dimensional middle representation, which is then concatenated with the quadrotor attitude (quaternion) and forward velocity (scalar) before feeding into either LSTM or fully-connected layers. With the exception of the ConvNet model, all models are roughly $\sim$3M parameters (Table \ref{table:model_specs}). This model size ensures fast enough computation for in-flight processing in a robotics control loop (30Hz on the hardware platform presented in Section \ref{subsec:hardware-setup}). Certain models such as a fully transformer-based architecture were attempted, but performed poorly and are not presented (unless a fully-connected head was used, as is described below as ViT). Architecture details are online \cite{bhattacharya2024vision}.\\
\indent\textit{ViT.} We utilize a transformer based encoder inspired by Segformer \cite{segformer}, and apply two transformer blocks in hierarchical fashion, followed by upsampling using pixel shuffle \cite{pxshuffle}, and mixing of information across hierarchies using a convolution operation. This incorporates information at multiple scales. This ViT is followed by fully connected layers.\\
%
\indent\textit{ViT+LSTM.} This is identical to the \textit{ViT} model but includes a multi-layer LSTM before the fully connected layers.\\
\indent\textit{ConvNet.} A lightweight CNN serving as a reference model resembling the first, and still widely used, successful deep learning architecture for object detection or end-to-end vision-based control for robots. This model is the only learned baseline significantly smaller than 3M parameters. ResNet-18 \cite{he2016resnet}, a large CNN model that learns residuals of desired layer outputs from inputs, was also trained but found to perform similarly to the ConvNet model and worse than UNet model variations (below), demonstrating that simply increasing the number of parameters in the ConvNet does not lead to better performance.\\
\indent\textit{LSTMnet.} A multi-layer LSTM module placed between the convolutional and fully connected layers is included to encourage temporal consistency in velocity predictions.\\
\indent\textit{UNet+LSTM.} The symmetric encoder-decoder architecture popular in segmentation tasks includes skip connections that incorporate local and global information and help prevent vanishing gradients. The reduction to a lower-dimensional feature space also may improve generalizability to unseen environments. A multi-layer LSTM follows the UNet. Our model includes multiple skip connections to improve performance and form a strong comparison.\\

\input{tables/table-model-params}


\subsection{Hardware setup}
\label{subsec:hardware-setup}


We use the Falcon250 custom quadrotor platform \cite{tao2023seer} (Figure \ref{fig:falcon250}) running the \texttt{kr\_mav\_control} \cite{githubGitHubKumarRoboticskr_mav_control} open-source quadrotor control stack. All experiments are performed in a motion capture arena for accurate state estimate. The Intel Realsense D435 camera utilizes structured infrared light and stereo matching to calculate depth. These images are resized and invalid depth pixels are set to a fixed value; model inference is done on the Intel NUC 10 onboard computer (i7-10710U CPU). The velocity command prediction is sent to the control stack, which outputs a SO(3) command (orientation and thrust) that is sent to the open-source \texttt{PX4} controller \cite{meier2015px4}. While we train models on floating spherical objects in simulation, we test with free-standing cylinders and blocks in hardware experiments.

%% file: tables/table-model-params.tex
\begin{table}
    \centering
    \begin{tabular}{lccc}
        \cline{1-4}
        \textbf{Model} & \textbf{Parameters} & \textbf{CPU (ms)} & \textbf{GPU (ms)} \\ \cline{1-4}\vspace{-7pt}\\
        ConvNet & 235,269 & 0.2 & 0.4 \\
        LSTMnet & 2,949,937 & 4.8 & 0.7 \\
        UNet+LSTM & 2,955,822 & 5.6 & 1.7 \\
        ViT & 3,101,199 & 5.6 & 1.6 \\
        ViT+LSTM & 3,563,663 & 9.2 & 3.8 \\ \cline{1-4}
    \end{tabular}
    \caption{Model size and inference times per model. Times are presented from a 12th Gen Intel i7-12700H CPU and Nvidia GeForce RTX 3060 GPU, with benchmark times obtained using TorchBench \cite{torchbench} via 1000 single-threaded iterations.}
    \label{table:model_specs}
\end{table}

%% file: results.tex
\subsection{Collision metrics}
\label{subsec:metrics}

\begin{figure}
    \centering
    \begin{subfigure}[t]{.47\linewidth}
        \centering
        \includegraphics[width=1.0\linewidth]{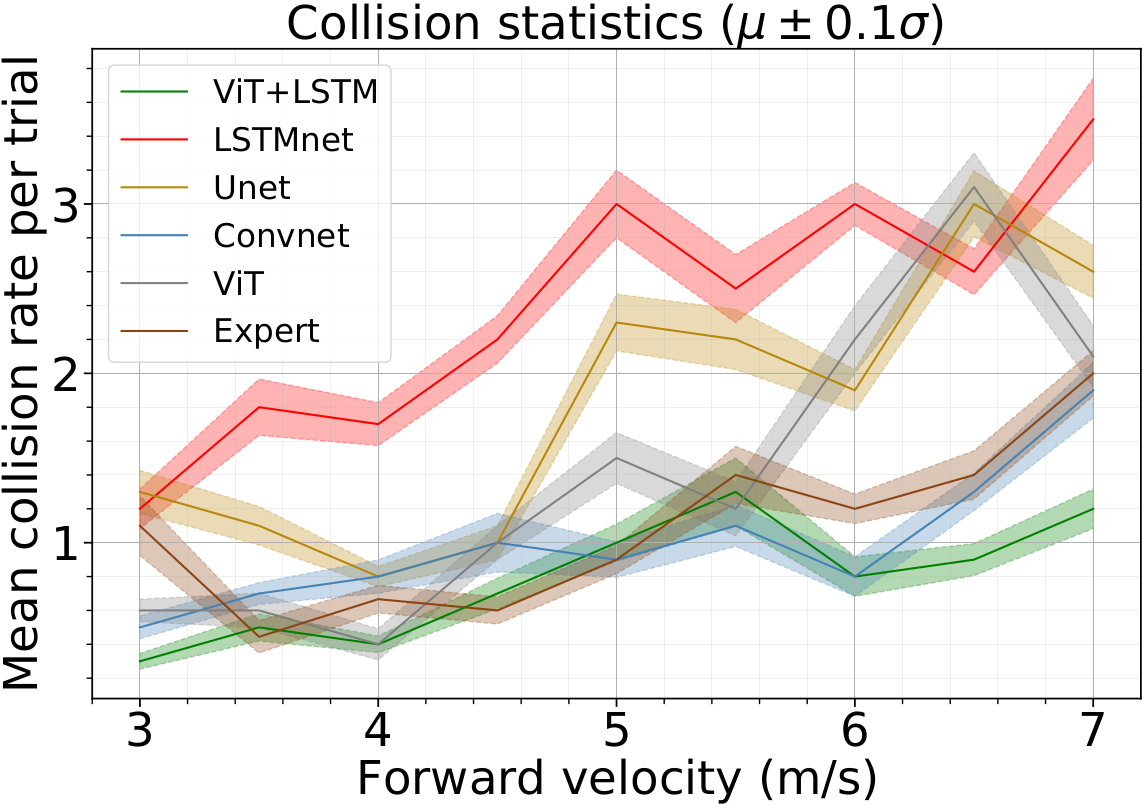}
        \vspace{-18pt}\caption{Spheres collision rate.}
        \label{fig:spheres-col-rate}
    \end{subfigure}%
    ~
    \begin{subfigure}[t]{.47\linewidth}
        \centering
        \includegraphics[width=1.0\linewidth]{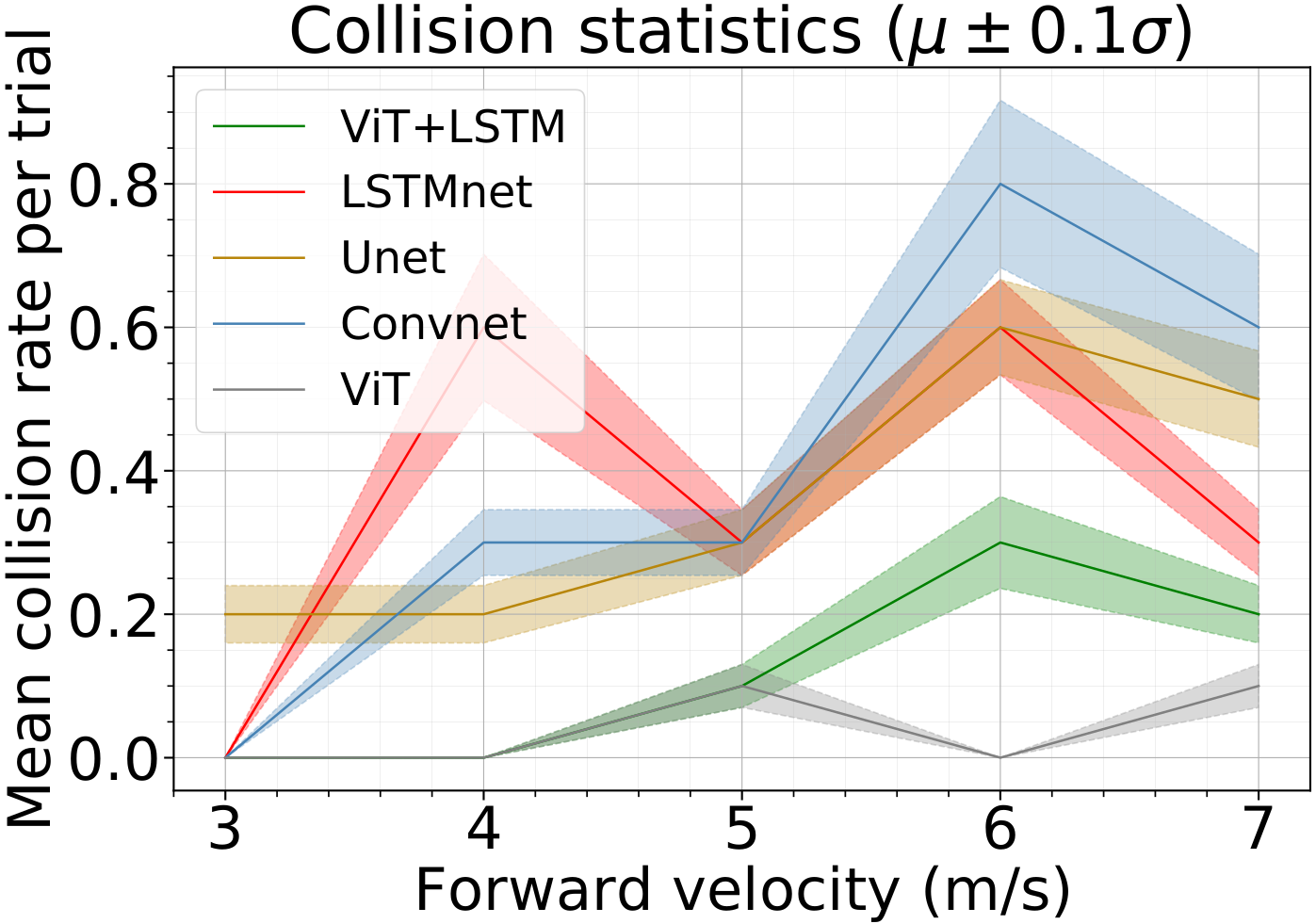}
        \vspace{-18pt}\caption{Trees collision rate.}
        \label{fig:trees-col-rate}
    \end{subfigure}
    \vfill
    \begin{subfigure}[t]{.43\linewidth}
        \centering
        \includegraphics[width=1.0\linewidth]{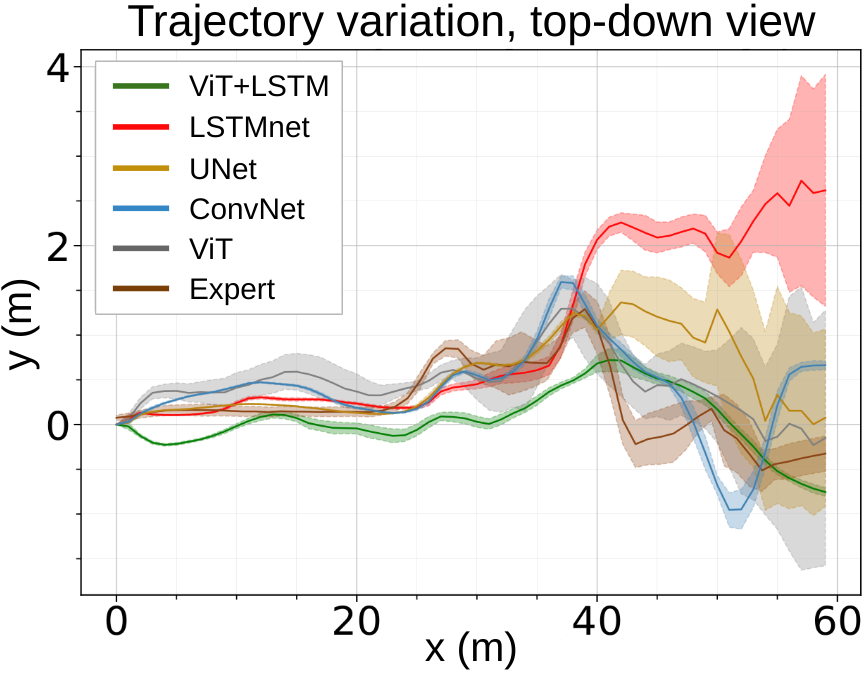}
        \vspace{-18pt}\caption{Top-down view of paths.}
        \label{fig:traj-vis}
    \end{subfigure}%
    ~
    \begin{subfigure}[t]{.55\linewidth}
        \centering
        \includegraphics[width=1.0\linewidth]{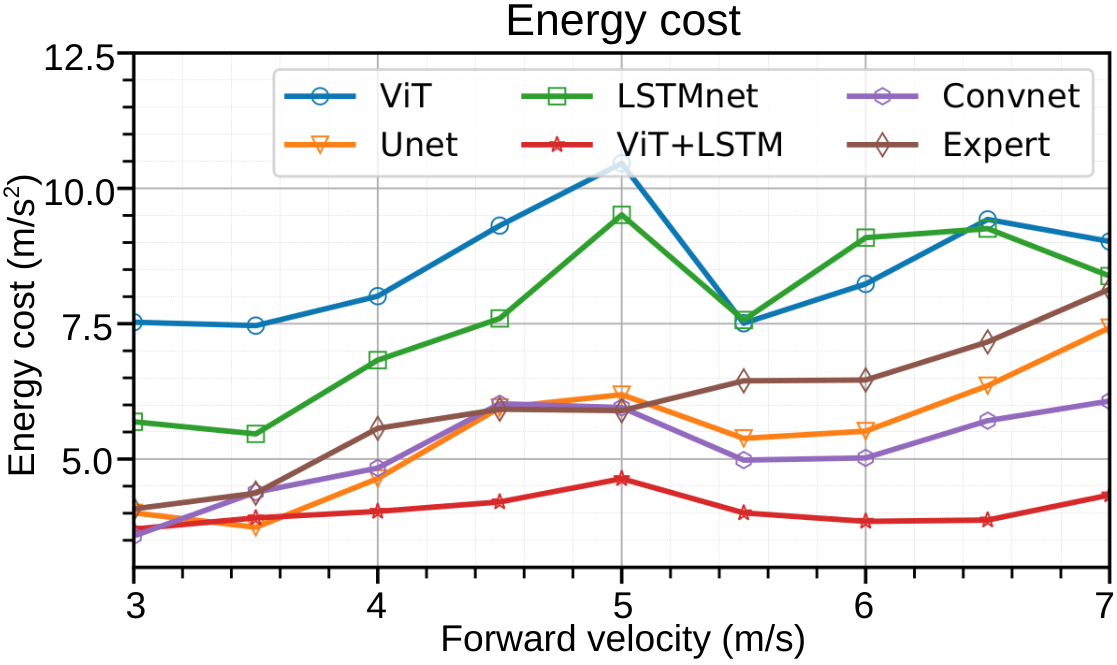}
        \vspace{-18pt}\caption{Estimated energy cost.}
        \label{fig:spheres-energy}
    \end{subfigure}
    \caption{In \ref*{fig:spheres-col-rate}, collision rates in a previously-unseen Spheres environment are lower for ViT+LSTM (green) versus the expert and other models beyond 5m/s. \ref*{fig:trees-col-rate} shows that in a novel-obstacle Trees environment, the ViT-based models (green, grey) generalize and perform substantially better than other models. \ref*{fig:traj-vis} qualitatively depicts the path distribution for each model in a fixed scene, where ViT+LSTM (green) appears to consistently take a more direct path through the cluttered environment (note non-equal axis aspect ratios). \ref*{fig:spheres-energy} presents the estimated energy cost, where ViT+LSTM is better than either component used alone.}
    \label{fig:sim-results}
\end{figure}

We present collision rates in Figure \ref{fig:sim-results} for two simulation environments; Spheres (\ref{fig:spheres-col-rate}) has obstacles similar to those seen during training, and Trees (\ref{fig:trees-col-rate} contains trees that are previously unseen to the models. Mean collision rates are calculated over multiple trials performed in varying obstacle configurations, with velocities of 3-7m/s and 90 trials for each model in the Spheres environment, and 50 trials each in the Trees environment. As supported by prior work, collision rates increase as forward velocity increases. However, ViT+LSTM outperforms other models as flight speed increases beyond 6m/s, and is the only model to consistently outperform the expert beyond 5m/s. This is likely due to the recurrent nature of the ViT+LSTM, in contrast to the privileged expert which independently calculates a desired waypoint and corresponding velocity at each re-planning timestep. Notably, models only composed of the ViT+LSTM components, namely the ViT and LSTMnet models, present worse collision rates than ViT+LSTM as speeds reach 7m/s. Success rate (i.e., zero-collision rate) plots for both environments can be found in the supplementary material of this paper's online version \cite{bhattacharya2024vision}.

Mean time spent in collision per obstacle, per trial was also used to analyze model behavior (a plot is presented in online material \cite{bhattacharya2024vision}). Since any collisions during expert rollouts causes the collision-free waypoint search to fail, briefly stalling the quadrotor planner, the expert presents large values for this metric over multiple trials (1.20s at 3m/s, 0.45s at 7m/s). However, all end-to-end learned models exhibit much lower values at this metric (worst values of 0.30s at 3m/s, 0.16s at 7m/s), demonstrating that they do not learn this stalling behavior, and again the ViT+LSTM model outperforms all other models at speeds of 4.5m/s and greater.

\subsection{Path and command characteristics}
\label{subsec:characteristics}


The trajectory paths in Figure \ref{fig:traj-vis} exhibit the path distribution of each model during 60m of forward flight through a fixed Spheres environment. 
The ConvNet and ViT+LSTM demonstrate significantly less variance in paths through the scene, while
the ViT model exhibits changing variance in the chosen path as it progresses towards the goal. Notably, the ViT+LSTM takes the most direct path through the obstacles, with a maximum mean lateral deviation of 0.75m from the starting $y=0$ position.

Figure \ref{fig:spheres-energy} presents the calculated energy cost as described in \cite{yu2023avoidbench}, which for real drones translates to longer flight time with a limited battery life (lower energy cost is better). The ViT model has a drastically improved energy cost with the addition of recurrence (ViT+LSTM). And interestingly, while all models and the expert policy present worsening energy cost as forward velocity increases, ViT+LSTM does not increase by an appreciable amount. Acceleration and energy cost plots for both Spheres and Trees are shown in the online material \cite{bhattacharya2024vision}.

\subsection{Network feature analysis}
\label{subsec:nn-feature-analysis}

\begin{figure}
    \centering
    \begin{subfigure}[t]{.32\linewidth}
        \centering
        \includegraphics[width=1.0\linewidth]{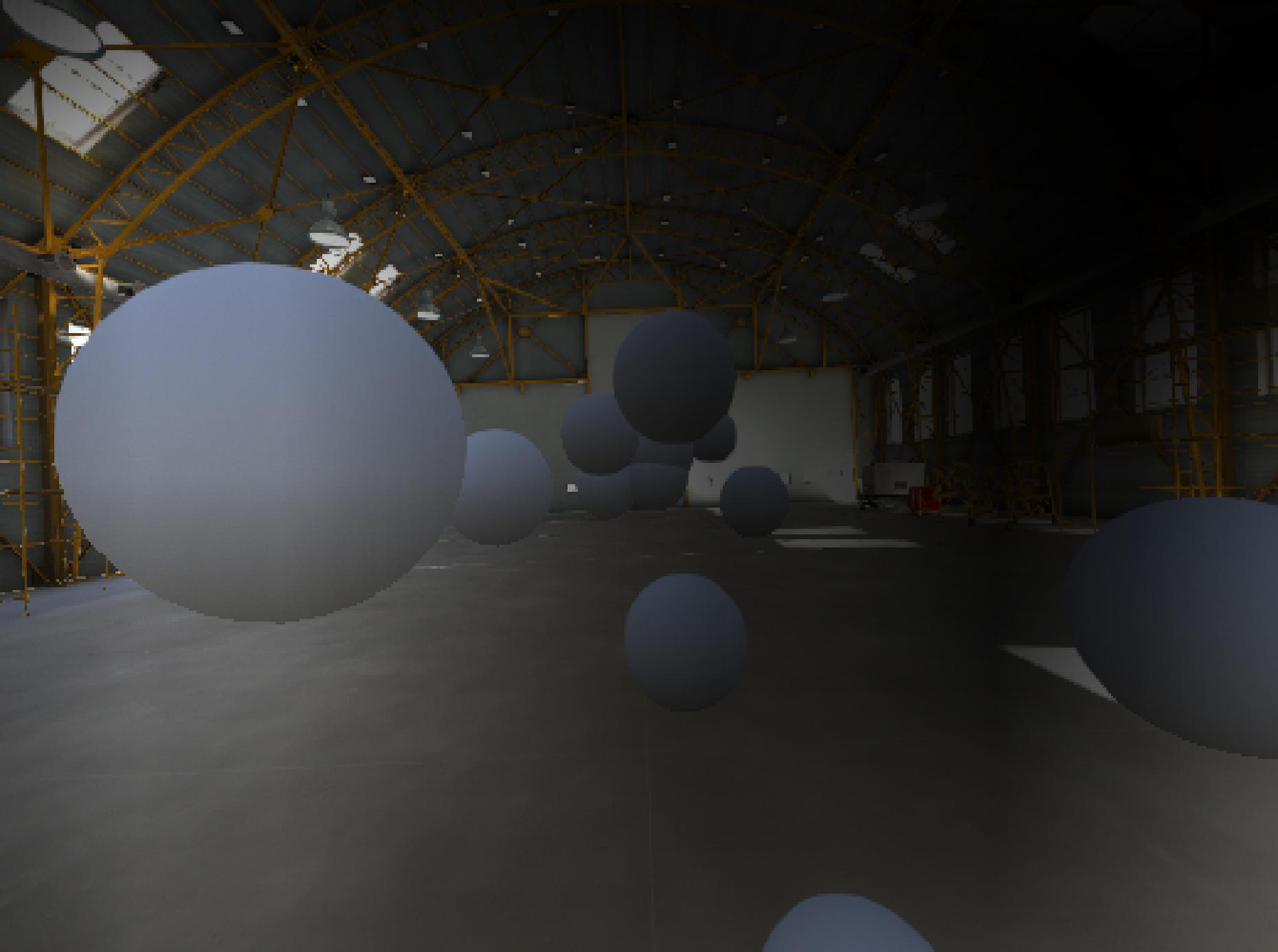}
        \vspace{-15pt}
        \caption{Sim ConvNet.}
        \label{fig:attmask-conv}
    \end{subfigure}%
    ~
    \begin{subfigure}[t]{.32\linewidth}
        \centering
        \includegraphics[width=1.0\linewidth]{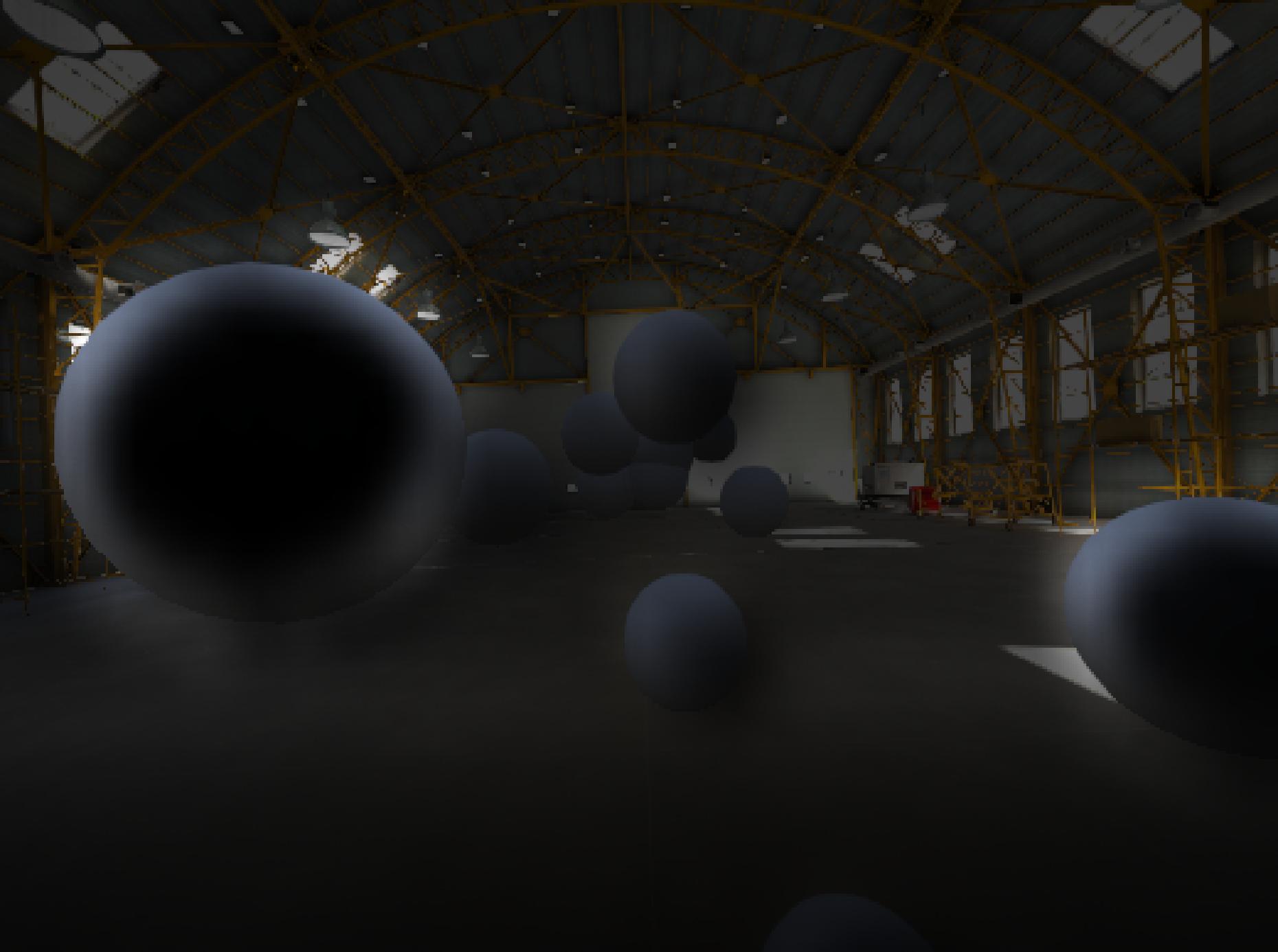}
        \vspace{-15pt}
        \caption{Simulated UNet.}
        \label{fig:attmask-unet}
    \end{subfigure}%
    ~
    \begin{subfigure}[t]{.32\linewidth}
        \centering
        \includegraphics[width=1.0\linewidth]{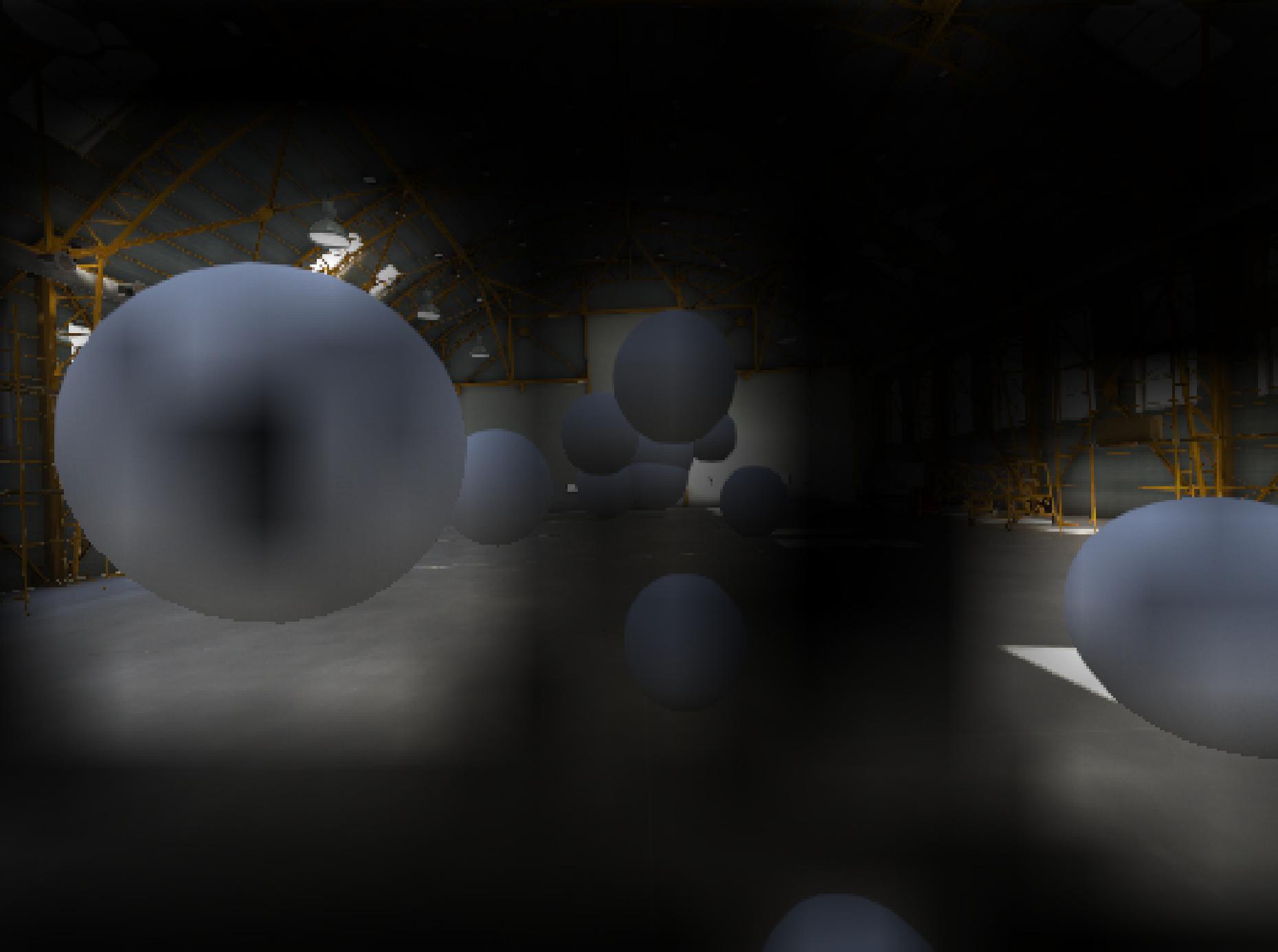}
        \vspace{-15pt}
        \caption{Simulated ViT.}
        \label{fig:attmask-vit}
    \end{subfigure}
    \vfill
    \begin{subfigure}[t]{.32\linewidth}
        \centering
        \includegraphics[width=1.0\linewidth]{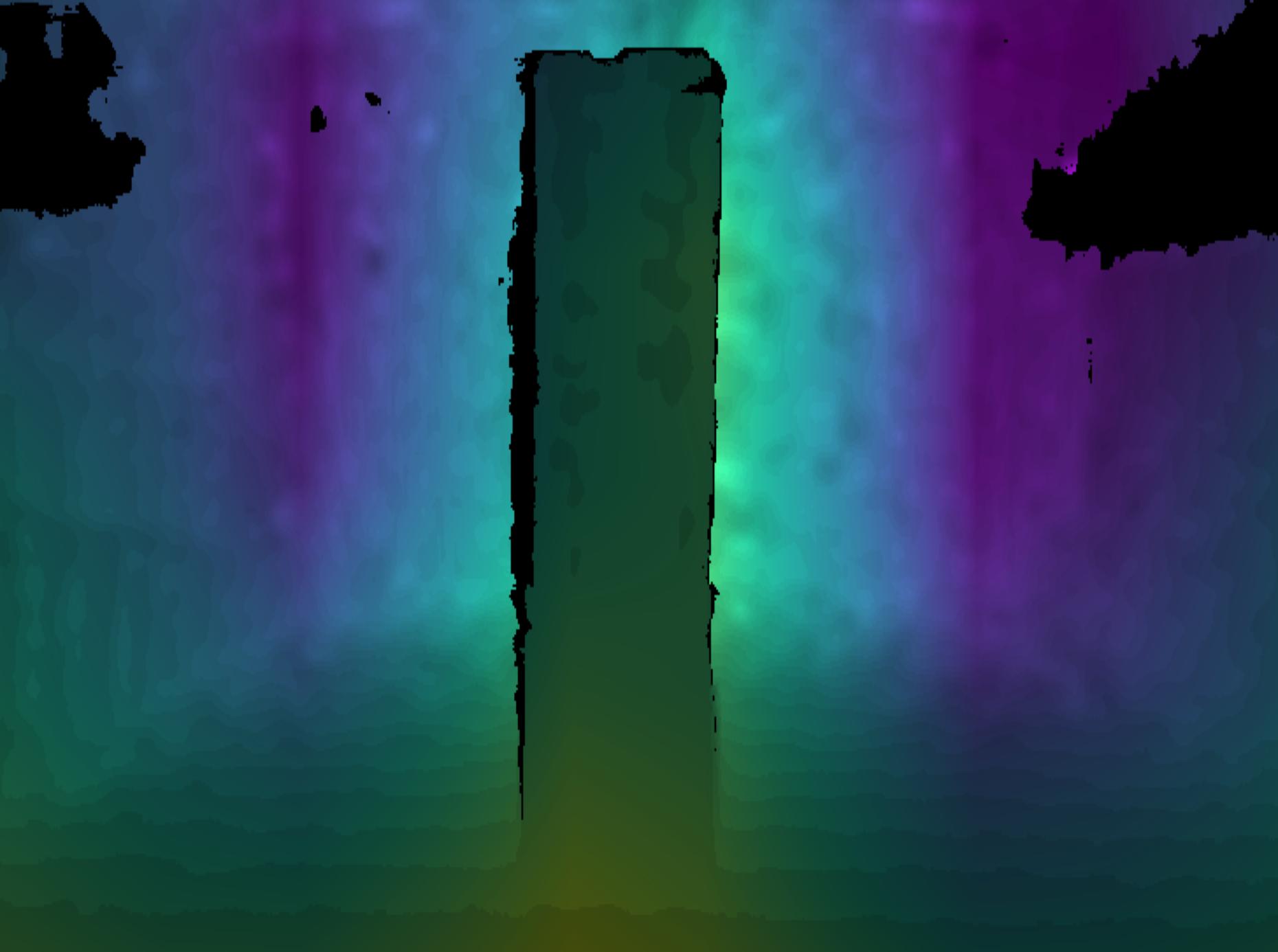}
        \vspace{-15pt}
        \caption{Real ConvNet.}
        \label{fig:attmask-real-conv}
    \end{subfigure}%
    ~
    \begin{subfigure}[t]{.32\linewidth}
        \centering
        \includegraphics[width=1.0\linewidth]{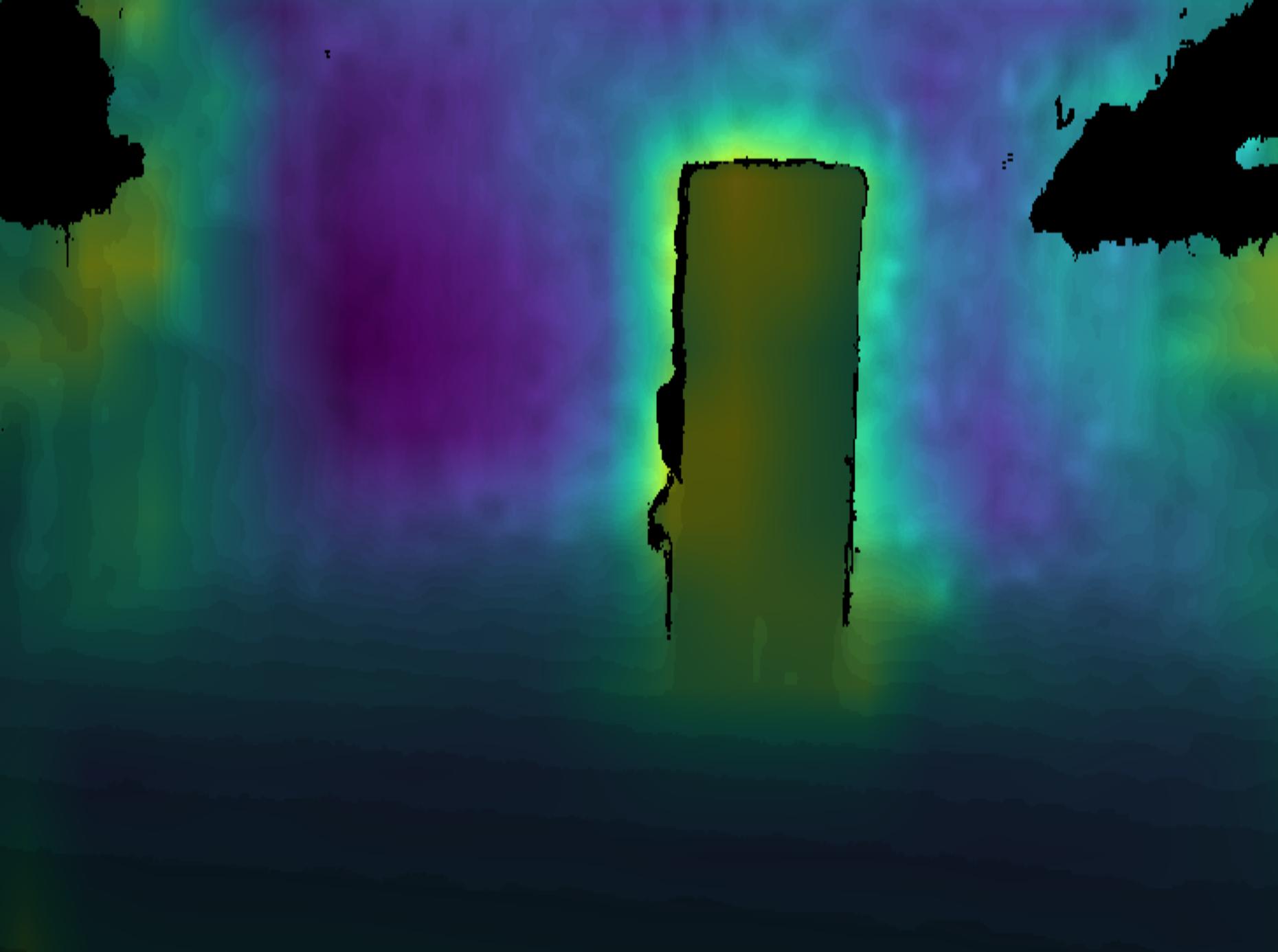}
        \vspace{-15pt}
        \caption{Real ViT.}
        \label{fig:attmask-real-vit}
    \end{subfigure}
    \caption{Attention map visualizations for each model's characteristic operation (sequential convolution, UNet-style convolution, or attention). Depth images are used as model input, but color images with light/dark or green/purple highlights (indicating high/low attention), are shown on color or depth images for simulated or real experiments, respectively.}
    \label{fig:attmasks}
\end{figure}

Figure \ref{fig:attmasks} highlights areas of the input, for simulated and real experiments, that return the strongest signal by the convolutional layers in ConvNet and UNet (via spatial activation maps), and by the attention layers in ViT (via attention maps). ConvNet highlights full obstacles with little specificity to their shape; UNet specifically highlights edges and disregards surrounding areas; the ViT model seems to both capture obstacle edges and surrounding context, thereby capturing nearby obstacles more than UNet, likely a result of the known behavior of ViTs to learn relationships between parts of an image \cite{vits}. In real experiments (Figures \ref{fig:attmask-real-conv}, \ref{fig:attmask-real-vit}) we find that the ConvNet attention map is more diffuse on the obstacle while ViT is much more specific.

\subsection{Generalizability}
\label{subsec:sim-generalizability}

To test model generalization capabilities, we zero-shot deploy the models in a simulation environment containing realistic tree models placed randomly in the scene (``Trees" environment). We present collision rates in Figure \ref{fig:trees-col-rate} with additional metrics shown in the online version of the paper \cite{bhattacharya2024vision}. We see that ViT-based models outperform all others, generalizing well. We further find that ViT-based models are the only models capable of generalization to a fly-through-window trial (often called a ``narrow-gap" trial), where only a tight distribution of viable collision-free paths exist from start to goal, compared to many such viable paths in the Spheres or Trees environments.

\subsection{Comparison with modular baselines}
\label{subsec:modular-baselines}

\begin{figure}
    \centering
    \begin{subfigure}[t]{.44\linewidth}
        \centering
        \includegraphics[width=1.0\linewidth]{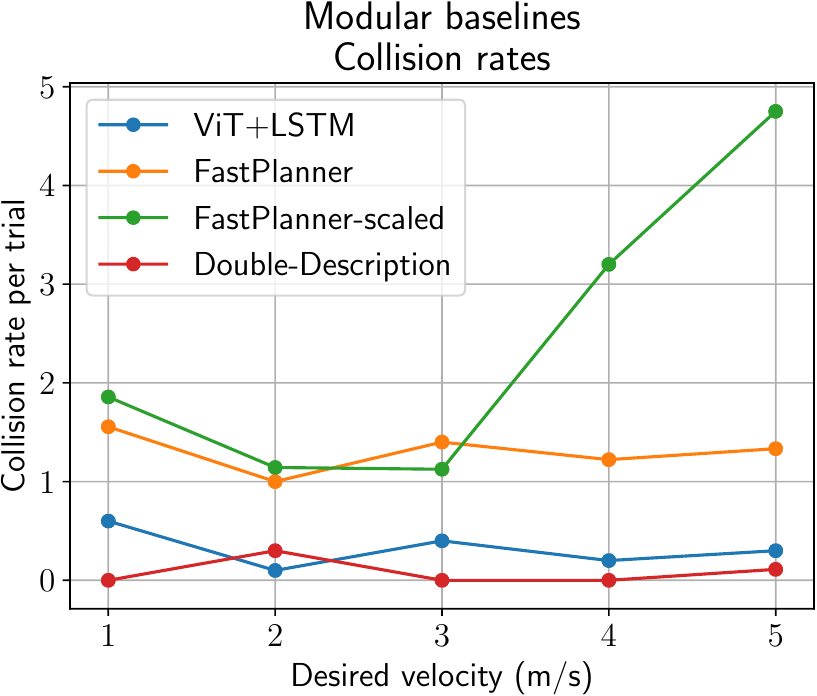}
        \vspace{-18pt}\caption{Collision rates.}
        \label{fig:modbaselines-colrate}
    \end{subfigure}%
    ~
    \begin{subfigure}[t]{.44\linewidth}
        \centering
        \includegraphics[width=1.0\linewidth]{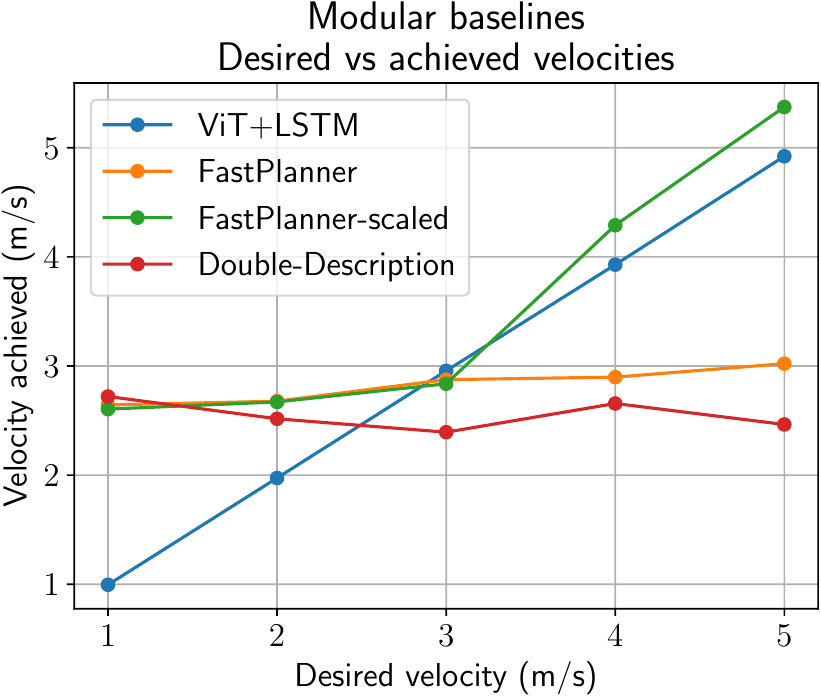}
        \vspace{-18pt}\caption{Velocity matching.}
        \label{fig:modbaselines-actualvel}
    \end{subfigure}
    \caption{Modular baselines FastPlanner \cite{fastplanner} and Double Description \cite{doubledescription}. Modular methods are designed for specific speeds (e.g., 3m/s) and require modification to scale to different speeds, which may cause drops in performance. In contrast, our end-to-end methods scale well to the 1-5m/s range with consistently low obstacle collision rates.}
    \label{fig:modbaselines}
\end{figure}

We compare the proposed end-to-end ViT+LSTM to two state-of-the-art modular baselines in the simulated Spheres environment (Figure \ref{fig:modbaselines}). FastPlanner \cite{fastplanner, yuwei-wu-fastplanner-github} finds an optimal initial path given multiple depth image observations in a receding horizon fashion, and fits a B-spline that is iteratively optimized to guarantee dynamic feasibility and be non-conservative. However, as this method is designed for flight around 3m/s, we additionally create a FastPlanner-scaled method which appropriately scales the output velocity to be closer to the desired velocities in the range 1-5m/s. However, this forced decrease or increase in flight speed either causes the FastPlanner algorithm to time-out (for $<$3m/s only valid trials' metrics are presented) or not keep up with observations (causing increased collisions), similar to the FastPlanner baseline metrics reported in \cite{Loquercio_2021}. Double Description \cite{doubledescription, yuwei-wu-doubledescription-github} is similarly designed for 3m/s flight, but produces fewer collisions, on par with ViT+LSTM. In contrast to these modular baselines, we find that our end-to-end method is the only one that not only appropriately flies at any desired speed (Figure \ref{fig:modbaselines-actualvel}), but also scales better with changing speed by maintaining consistently-low collision rates compared to modular baseline methods (Figure \ref{fig:modbaselines-colrate}).

\subsection{Ablation against state information}
\label{subsec:ablations}

\input{tables/table-no-state-info}

We perform an ablation against providing quadrotor orientation and forward velocity to the models by re-training without this data; success rates are presented in Table \ref{tab:no-metadata-sr} for Spheres and Trees across three velocities. This ablation is particularly relevant to high-speed flight since unstructured or windy environments may cause noisy state estimates and variable velocity. We find that ViT+LSTM performed better \textit{with} the state information across slow and fast speeds and across environments, while ConvNet only benefited in the Spheres environment but performed substantially worse when tested in the unseen, generalization Trees environment, particularly as speeds increase.

\subsection{Real world experiments}
\label{subsec:hardware-demos}

\begin{figure}[h]
    \centering
    \begin{minipage}{.38\linewidth}
        \centering
        \begin{subfigure}{\linewidth}
            \centering
            \includegraphics[width=1.0\linewidth]{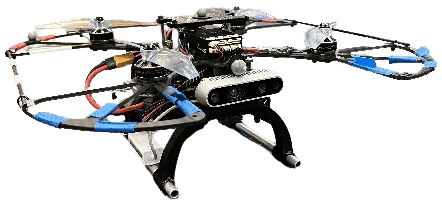}
            \vspace{-15pt}
            \caption{Falcon250.}
            \label{fig:falcon250}
        \end{subfigure}
        \vfill
        \begin{subfigure}{\linewidth}
            \centering
            \includegraphics[width=1.0\linewidth]{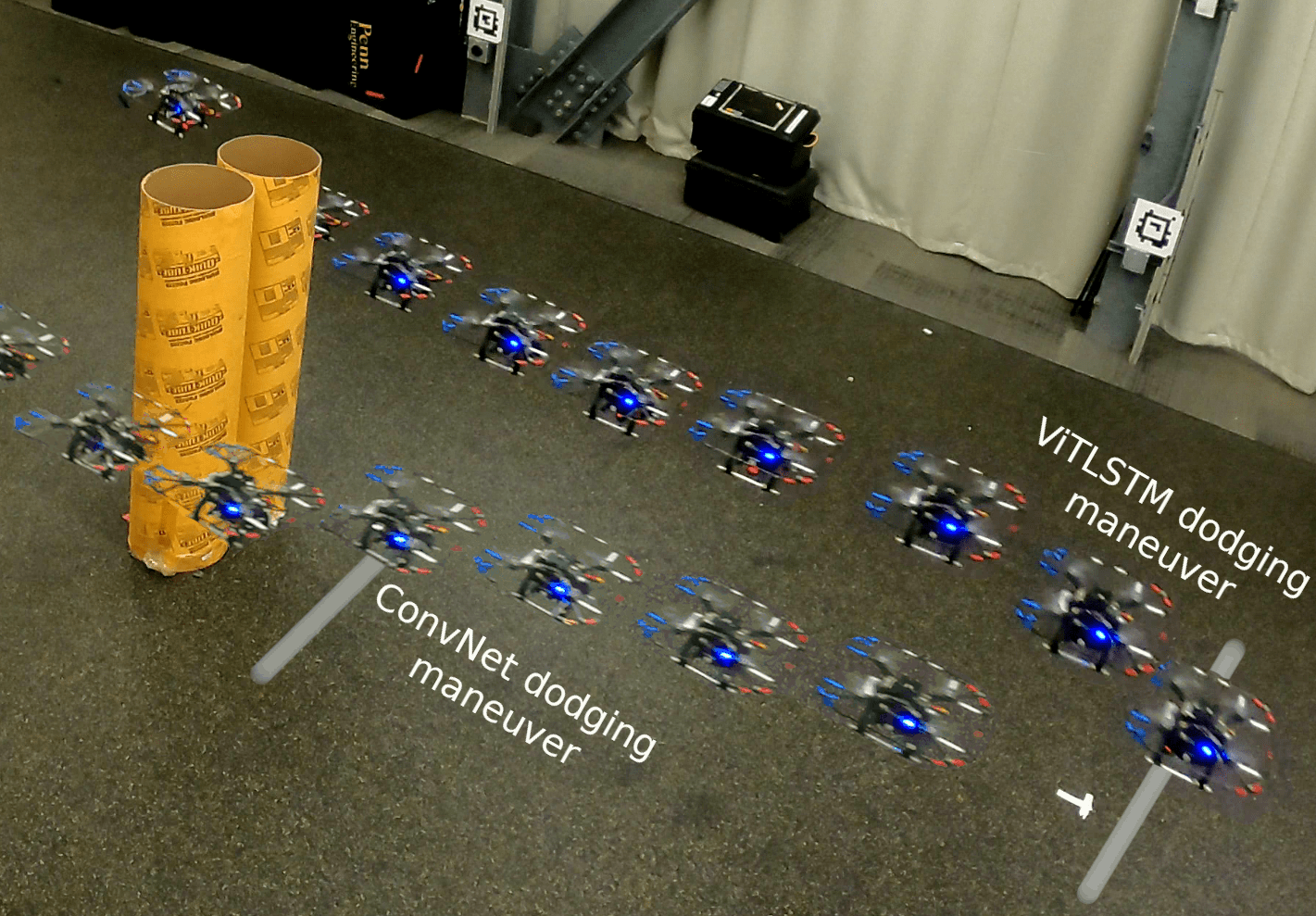}
            \vspace{-15pt}
            \caption{1m/s trials.}
            \label{fig:realexp-convnet-and-vitlstm}
        \end{subfigure}
    \end{minipage}%
    ~
    \begin{minipage}{.45\linewidth}
        \centering
        \begin{subfigure}{\linewidth}
            \centering
            \includegraphics[width=0.83\linewidth]{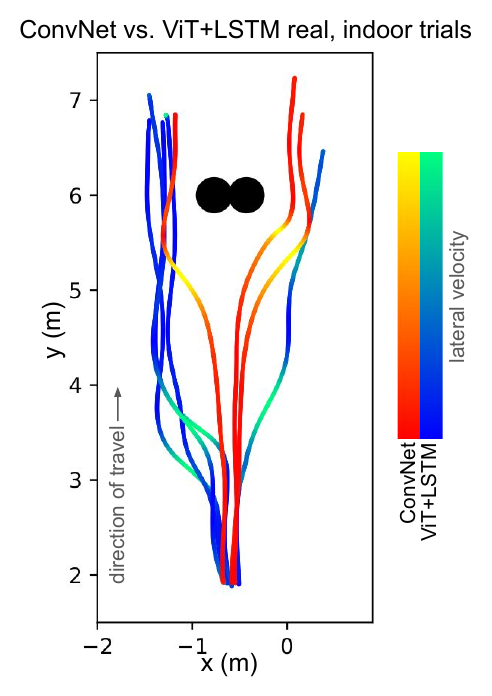}
            \vspace{-10pt}
            \caption{ConvNet vs. ViT+LSTM.}
            \label{fig:realexp-trajs}
        \end{subfigure}
    \end{minipage}
    \caption{Representative real trials using the Falcon250 (\ref*{fig:falcon250}) evaluating ConvNet and ViT+LSTM in the same single, rigid obstacle avoidance task (\ref*{fig:realexp-convnet-and-vitlstm}). Paths are plotted in \ref*{fig:realexp-trajs}, where lighter shades on the red (ConvNet) and blue (ViT+LSTM) trajectories represent higher lateral velocities. ViT+LSTM consistently avoids earlier than the ConvNet model.}
    \label{fig:real-experiments}
\end{figure}

We zero-shot deploy ViT+LSTM in single and multi-obstacle configurations requiring collision avoidance and flight-through-gap behaviors at high speeds up to 7m/s. The model runs inference onboard at 30Hz (CPU) during all trials. Figure \ref{fig:frontpage} shows representative examples of two such trials, with \ref{fig:frontpage-dbg-im} showing the depth image input to the model with a red arrow representing the model's corresponding output velocity command. The quadrotor avoids obstacles in the real world with no model re-training or extensive system tuning; while depth camera images exhibit artifacts not seen in the simulated training data, such as high-frequency patterns, varying background appearance, and propellers in-view (uncropped depth images viewable on project website), end-to-end learned models performed effectively in many scenes at speeds of 1-7m/s. Figure \ref{fig:real-experiments} additionally demonstrates real-world differences in collision avoidance between the ConvNet and ViT+LSTM models.

%% file: tables/table-no-state-info.tex
\begin{table}
\centering
\begin{tabular}{clcccccc}
    \multirow{4}{*}{} & \multirow{4}{*}{} & \multicolumn{6}{c}{\textbf{Velocity (m/s)}} \\
      &  & \multicolumn{2}{c|}{3} & \multicolumn{2}{c|}{5}       & \multicolumn{2}{c}{7} \\ \cline{3-8}\vspace{-7pt}\\
      &  & \multicolumn{6}{c}{Orientation and velocity included?} \\
      & \textbf{Models} & \multicolumn{1}{c}{\xmark} & \multicolumn{1}{c|}{\cmark} & \multicolumn{1}{c}{\xmark} & \multicolumn{1}{c|}{\cmark} & \multicolumn{1}{c}{\xmark} & \multicolumn{1}{c}{\cmark} \\ \cline{2-8}\vspace{-7pt}\\
    \multirow{5}{*}{\rotatebox{90}{Spheres}} & ConvNet    & 27  & \multicolumn{1}{c|}{\textbf{63}}  & 18 & \multicolumn{1}{c|}{\textbf{54}} & 10 & \textbf{36} \\
                      & LSTMnet    & 27  & \multicolumn{1}{c|}{\textbf{45}}  & \textbf{18}  & \multicolumn{1}{c|}{9} & 9  & 9 \\
                      & UNet+LSTM  & \textbf{36}  & \multicolumn{1}{c|}{27}  & \textbf{18}  & \multicolumn{1}{c|}{9} & \textbf{27}  & 9 \\
                      & ViT        & \textbf{81}  & \multicolumn{1}{c|}{54}  & 18 & \multicolumn{1}{c|}{\textbf{27}} & 18 & 18 \\
                      & ViT+LSTM   & 54  & \multicolumn{1}{c|}{\textbf{72}}  & 45 & \multicolumn{1}{c|}{45} & 9 & \textbf{36} \\ \cline{2-8}\vspace{-7pt}\\
    \multirow{5}{*}{\rotatebox{90}{\centering\begin{tabular}[c]{@{}l@{}}Trees\\(unseen)\end{tabular}}} & ConvNet    & 90 & \multicolumn{1}{c|}{\textbf{100}}  & \textbf{81} & \multicolumn{1}{c|}{63} & \textbf{60} & 42 \\
                      & LSTMnet    & 90  & \multicolumn{1}{c|}{90}  & 40 & \multicolumn{1}{c|}{\textbf{63}} & 36 & \textbf{63} \\
                      & UNet+LSTM  & \textbf{100}  & \multicolumn{1}{c|}{72} & 60 & \multicolumn{1}{c|}{63} & 45 & \textbf{54} \\
                      & ViT        & 90 & \multicolumn{1}{c|}{\textbf{100}}  & 81 & \multicolumn{1}{c|}{81} & 81 & 80 \\
                      & ViT+LSTM   & 90 & \multicolumn{1}{c|}{\textbf{100}}  & 81 & \multicolumn{1}{c|}{80} & 50 & \textbf{70} \\ \cline{2-8}
\end{tabular}
\caption{Ablation of providing quadrotor orientation and forward velocity to models, with success rates ($\%$) for each model across the Spheres and previously-unseen Trees environments. ViT+LSTM (in Spheres and Trees) and ConvNet (in Spheres) suffer the most from lack of this data. Non-bolded numbers in a section indicate little difference.}
\label{tab:no-metadata-sr}
\end{table}

%% file: conclusion.tex
We present the first use of a vision transformer for end-to-end quadrotor control from depth images, with a comprehensive discussion of collision rates and command characteristics in simulation, comparison to other learned architectures and traditional modular methods, as well as deployment in challenging real-world trials. We find that the attention-based models (ViT and ViT+LSTM) outperform other models at high speeds in in-distribution simulation environments, and greatly outperform other architectures across all speeds in generalization environments, even extending from obstacle avoidance in forward flight to a fly-through-window task where other architectures fail. The best model combines attention and recurrence (ViT+LSTM), and consistently presents low collision rates while also maintaining the lowest energy cost of all models at speeds ranging 3-7m/s. Ablations show that across various speeds and environments, the attention-recurrent model improves with the inclusion of quadrotor state information. We also found that, in contrast to the presented end-to-end architecture, modular baseline methods fail to maintain low collision rates when scaling to varying flight speeds. We provide additional experiments and visualizations, as well as all code, datasets, and pretrained weights on the project website (Section \ref{sec:introduction}) to enable further research of attention-based end-to-end robotics control.

%% file: supp-more-results.tex
\begin{figure}
  \centering
  \begin{subfigure}{0.32\linewidth}
    \includegraphics[width=1.0\linewidth]{figures/spheres-colrates.png}
    \vspace{-15pt}\caption{}
    \label{fig:spheres-collision-per-trial}
  \end{subfigure}
  \hfill
  \begin{subfigure}{0.32\linewidth}
    \includegraphics[width=1.0\linewidth]{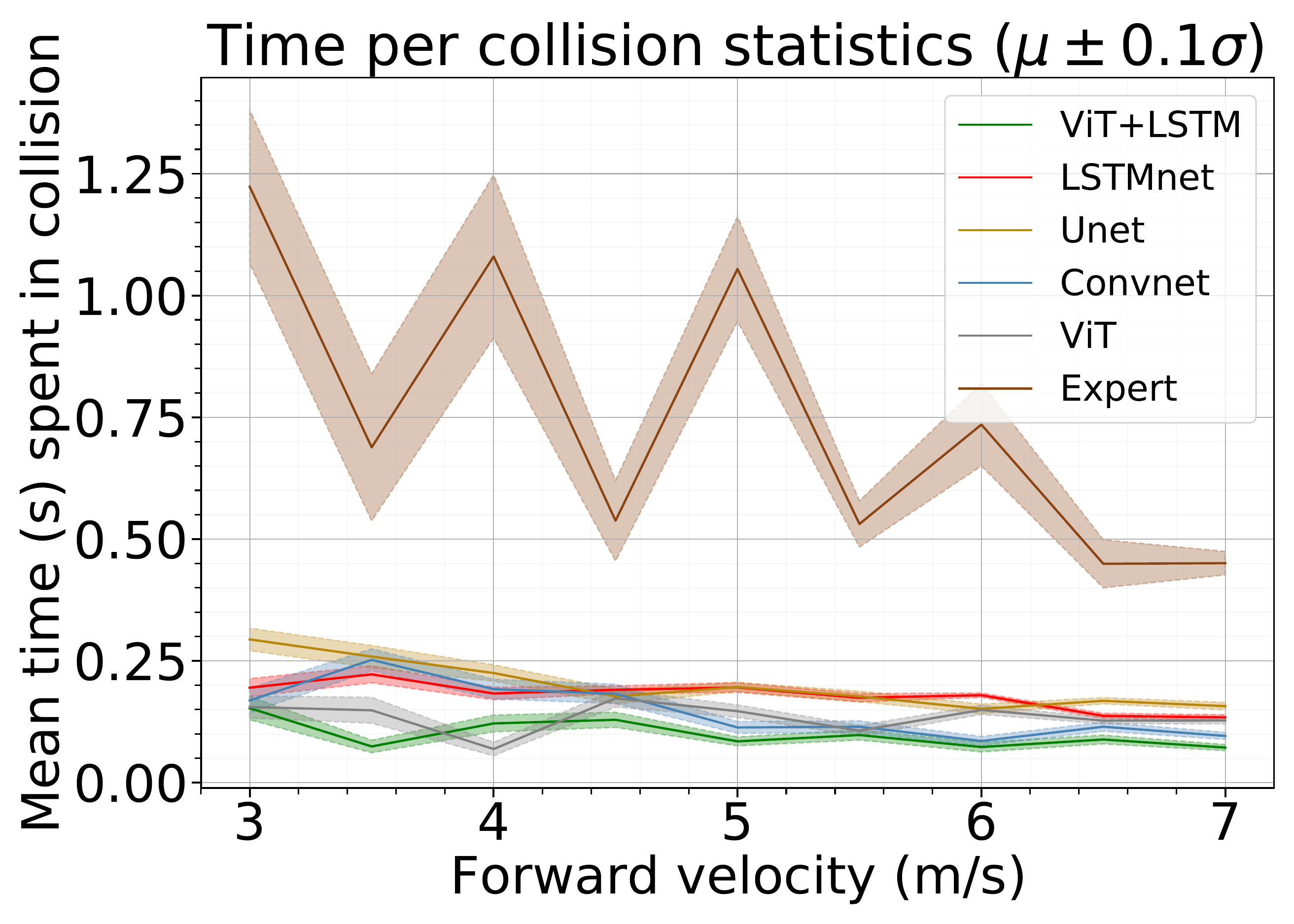}
    \vspace{-15pt}\caption{}
    \label{fig:spheres-collision-t-per-trial}
  \end{subfigure}
  \hfill
  \begin{subfigure}{0.32\linewidth}
    \includegraphics[width=1.0\linewidth]{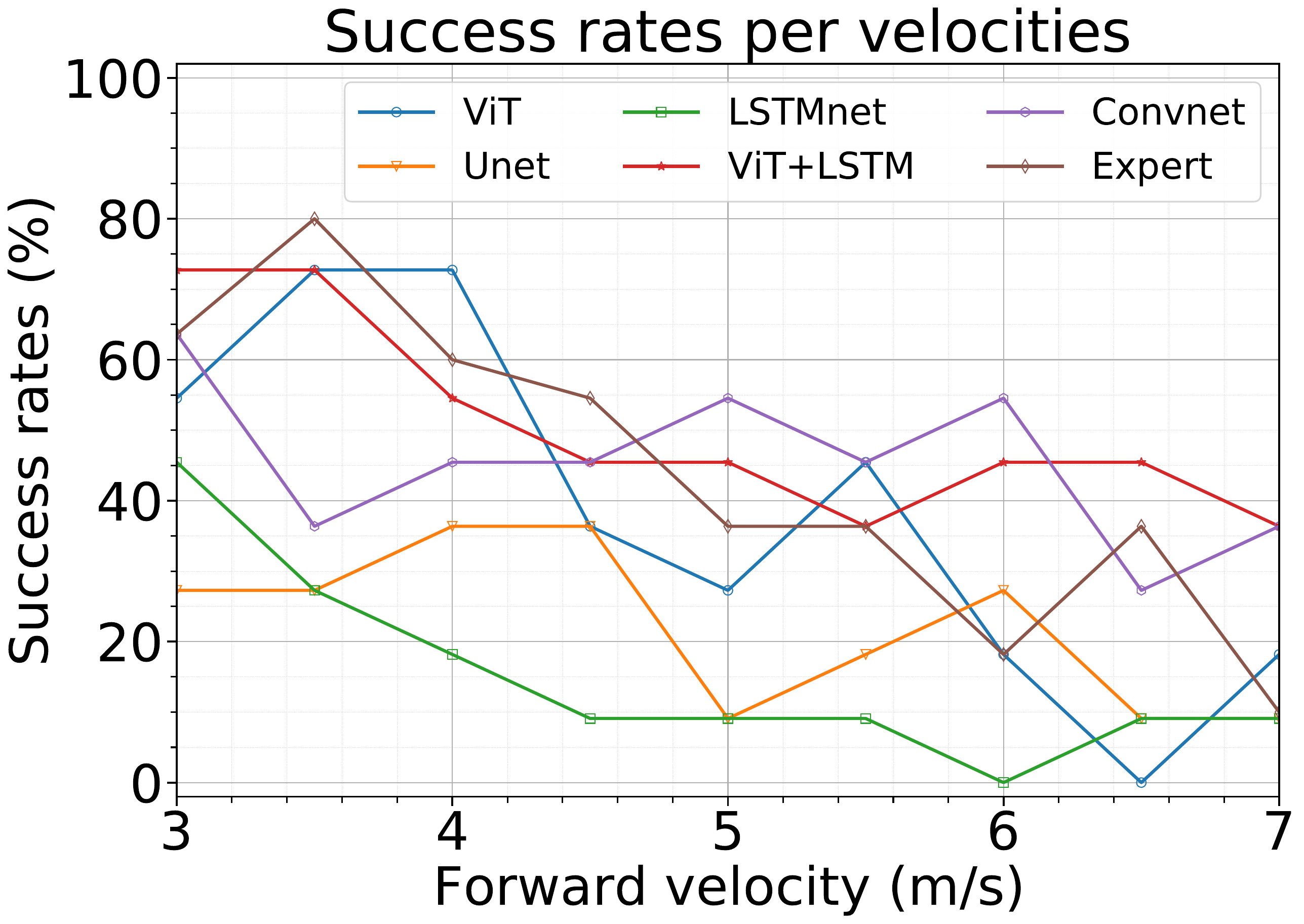}
    \vspace{-15pt}\caption{}
    \label{fig:spheres-sr}
  \end{subfigure}
  \caption{Collision and success metrics for each model, taken over 10 randomized, unseen simulation sphere configuration trials at each forward velocity. (a) Mean number of collisions per trial. Most models do not outperform the expert up to speeds of 5m/s, beyond which the ViT+LSTM model outperforms both the expert and other models. (b) Mean time spent in collision per obstacle, per trial, with each trial lasting between 8.5-20s. This metric tends to increase with collision ``severity", i.e. head-on collisions and large-obstacle collisions. The models outperform the expert, and ViT+LSTM does the best. (c) Success rates, i.e. the fraction of trials containing no collisions. ConvNet and ViT+LSTM are the most consistent models with forward velocity.}
  \label{fig:spheres-collision-metrics}
\end{figure}

Figure \ref{fig:spheres-collision-metrics} shows both collision metrics (collision rates per trial, and time spent in collision per obstacle, per trial) as discussed in the main text, as well as an additional success rate plot. It is expected that as forward flight speed increases (moving from left to right in each plot) the collision rate increases, time spent in collision decreases (due to the drone moving through obstacles faster), and success rate decreases.

\begin{figure}
  \centering
  \begin{subfigure}{0.37\linewidth}
    \includegraphics[width=1.0\linewidth]{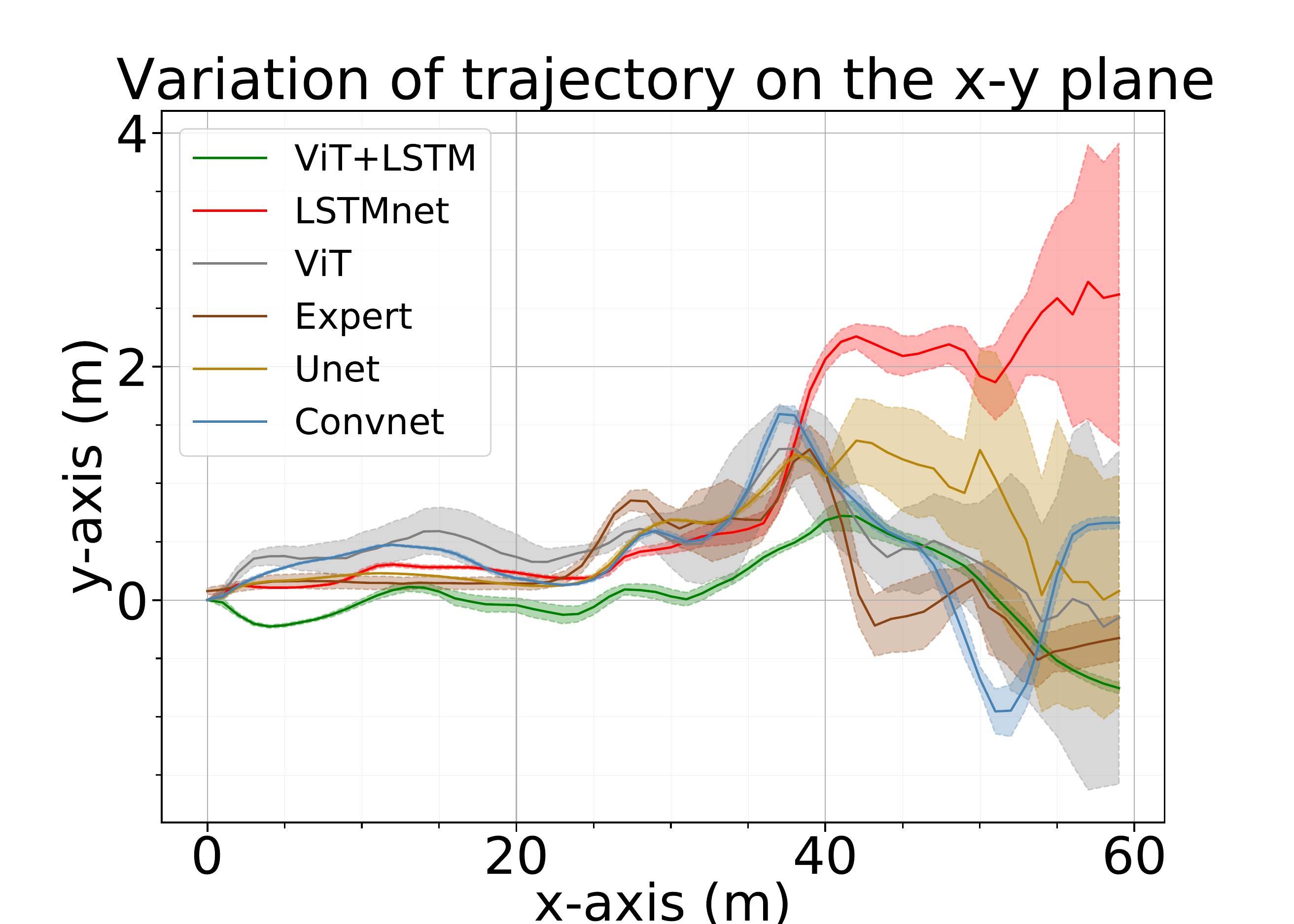}
    \vspace{-15pt}\caption{Top-down view.}
    \label{fig:traj-2d-topdown}
  \end{subfigure}
  \hfill
  \begin{subfigure}{0.37\linewidth}
    \includegraphics[width=1.0\linewidth]{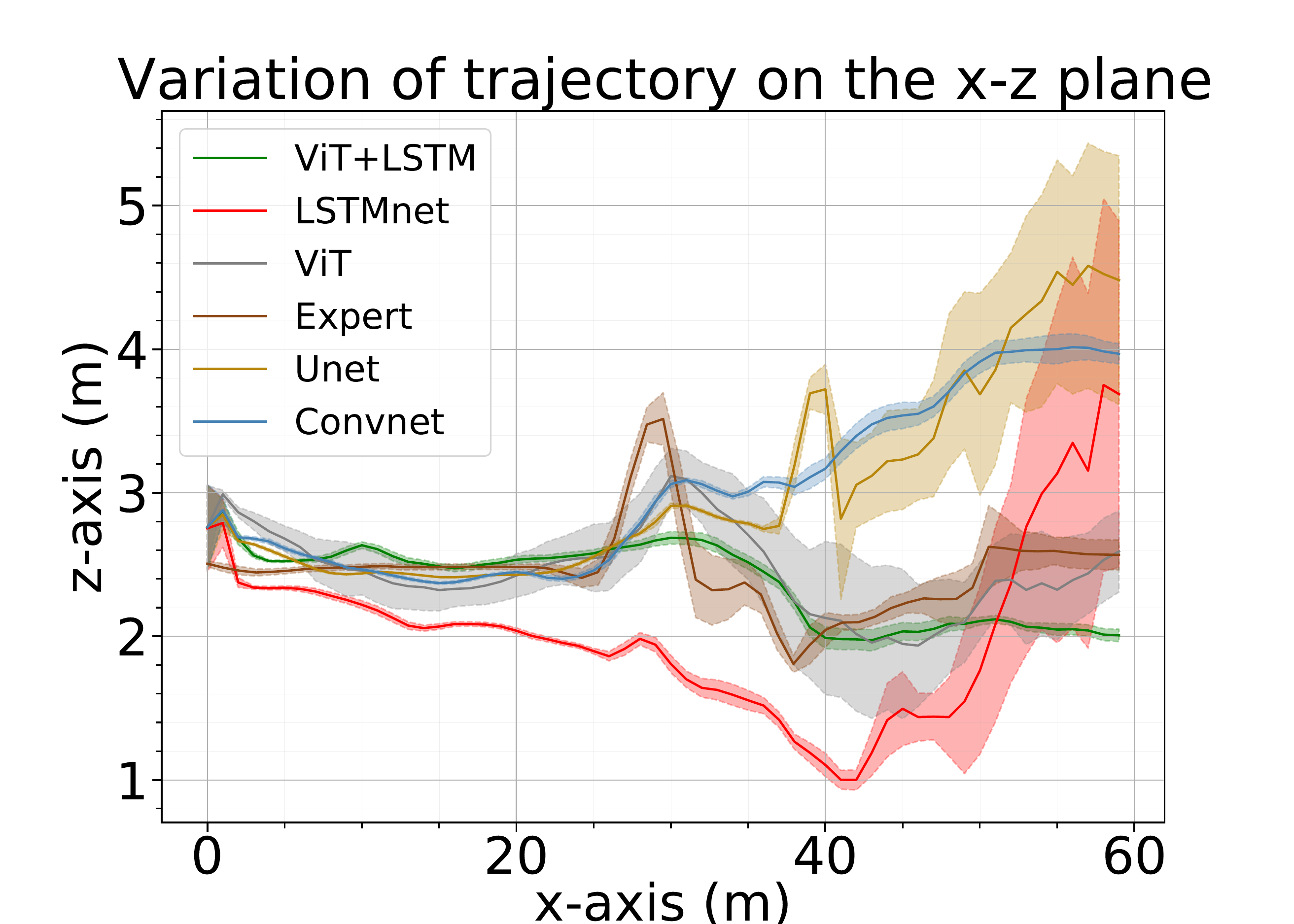}
    \vspace{-15pt}\caption{Side view.}
    \label{fig:traj-2d-sideview}
  \end{subfigure}
  \hfill
  \begin{subfigure}{0.23\linewidth}
  \centering
    \includegraphics[width=1.0\linewidth]{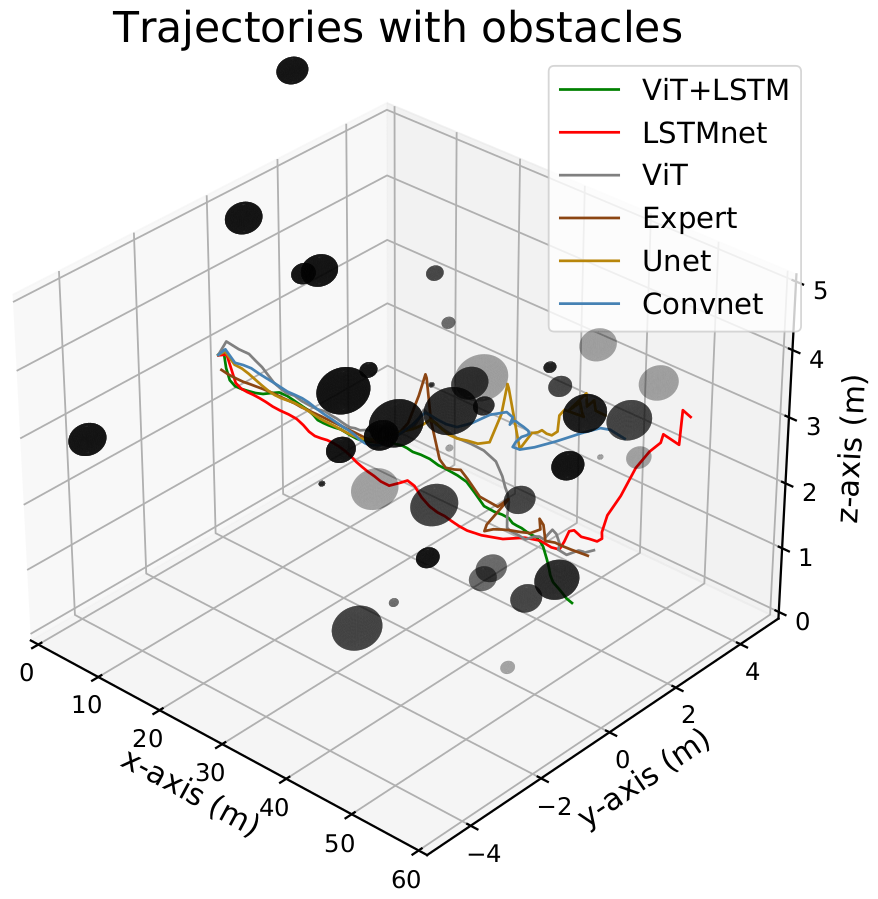}
    \vspace{-15pt}\caption{Persp. view.}
    \label{fig:traj-3d}
  \end{subfigure}
  \caption{Visualizing the trajectories taken by each model in a single test environment over multiple trials. Note that axes are displayed with non-equal aspect ratios. (a) Top-down view with $0.1\sigma$ distribution of the mean position, (b) side-view with $0.1\sigma$ distribution of the mean position, (c) 3D perspective of the mean paths.}
  \label{fig:traj-vis}
\end{figure}

Figure \ref{fig:traj-vis} shows the distribution of paths taken by each model through a fixed Spheres environment across multiple trials. As is pointed out in the main text, the ViT+LSTM takes the most direct path to the goal, with the least deviation from the $y=0,z=0$ reference straight-line trajectory while avoiding obstacles. We can also see it has less variance in its path over multiple trials compared to other models. Figure \ref{fig:traj-3d} shows the mean of each model's path distribution through the given obstacle field.

\begin{figure}
  \centering
  \begin{subfigure}{0.32\linewidth}
    \includegraphics[width=1.0\linewidth]{figures/trees-colrates.png}
    \vspace{-15pt}\caption{}
    \label{fig:trees-collision}
  \end{subfigure}
  \hfill
  \begin{subfigure}{0.32\linewidth}
    \includegraphics[width=1.0\linewidth]{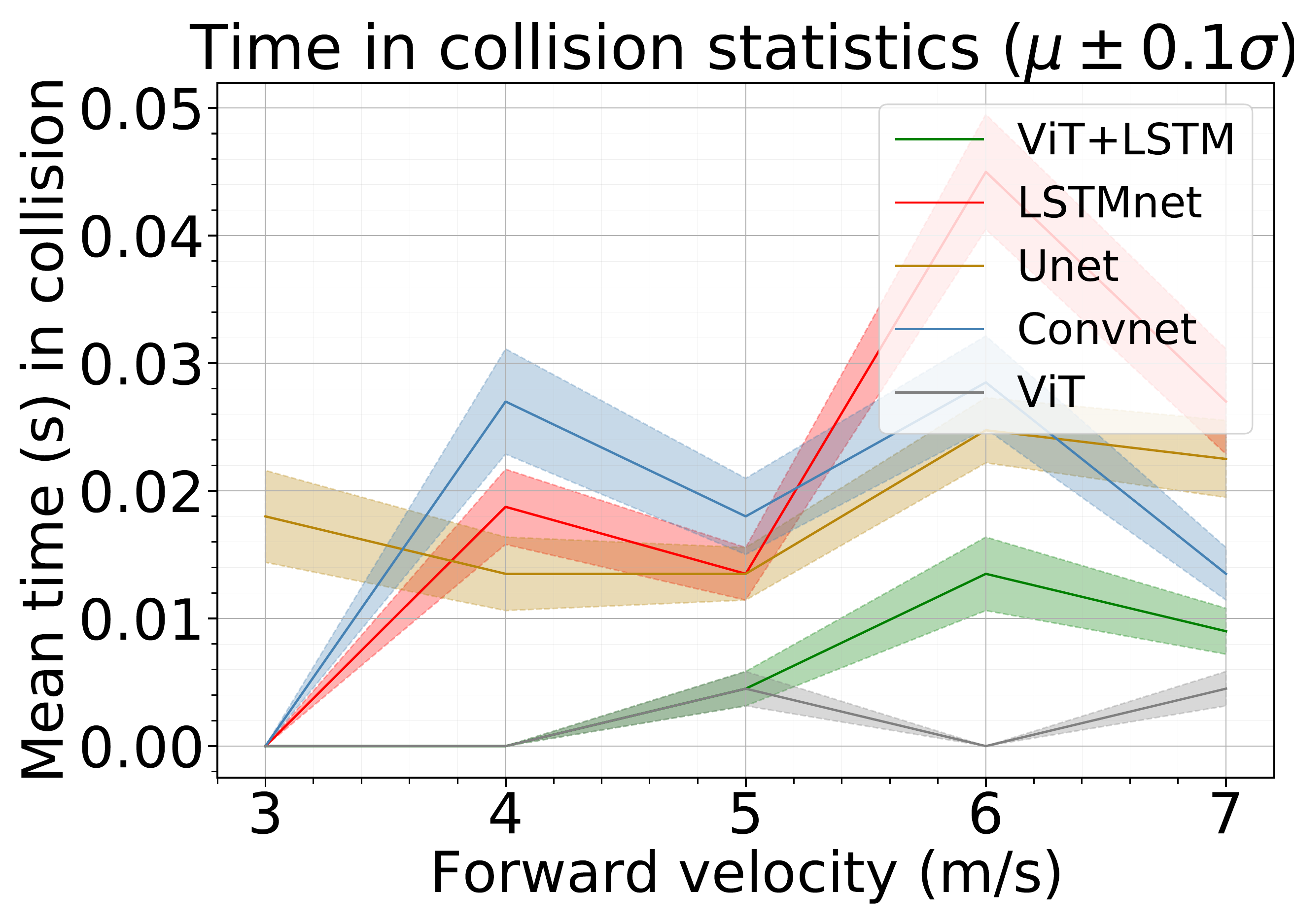}
    \vspace{-15pt}\caption{}
    \label{fig:trees-collision-t}
  \end{subfigure}
  \begin{subfigure}{0.32\linewidth}
    \includegraphics[width=1.0\linewidth]{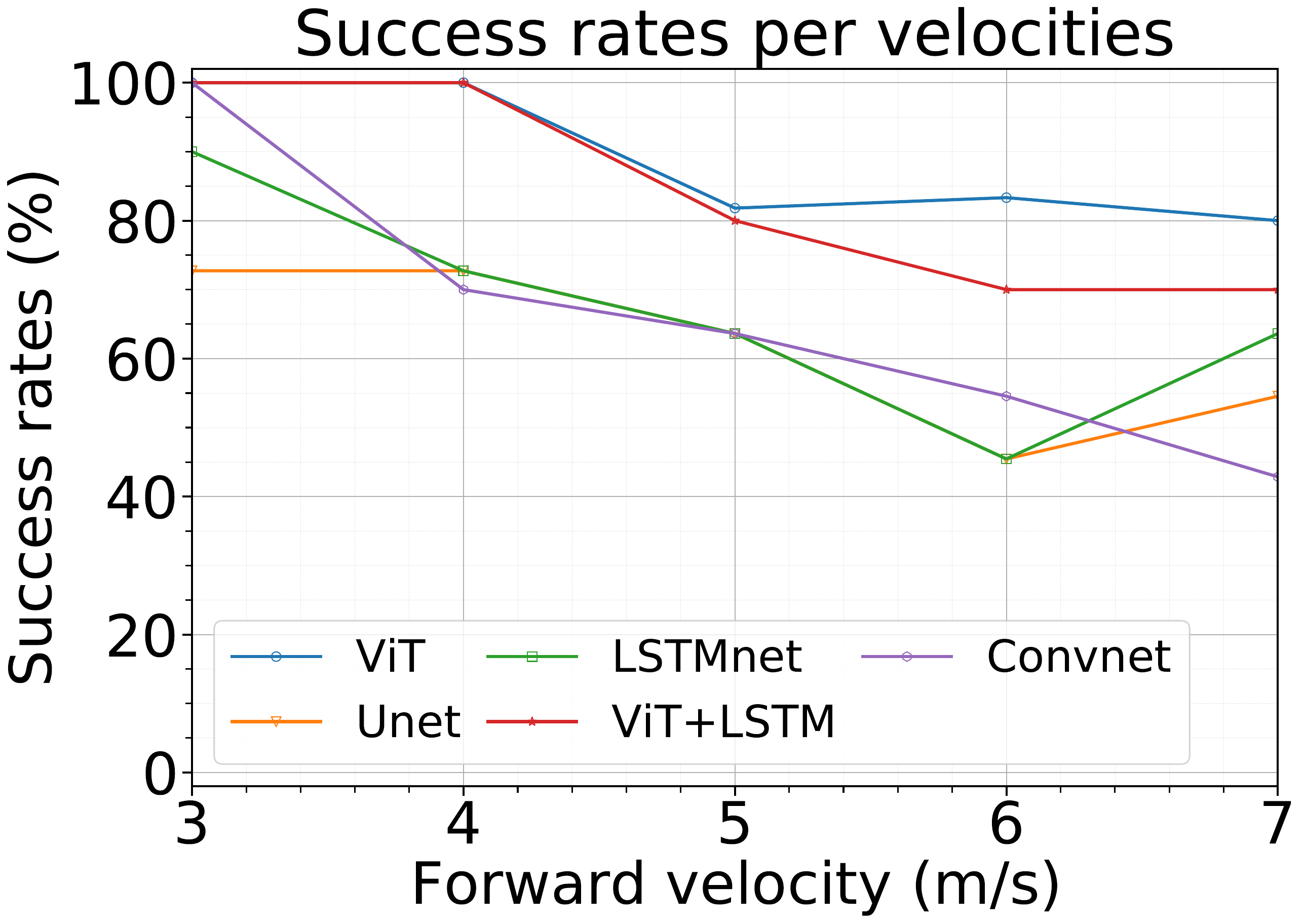}
    \vspace{-15pt}\caption{}
    \label{fig:trees-sr}
  \end{subfigure}
\caption{Generalization to Trees: collision and success metrics for each model, taken over 10 randomized tree environments. Since we did not train on Trees, we do not compare with the expert. (a) Mean number of collisions per trial. (b) Mean time spent in collision per obstacle, per trial. (c) Success rates, i.e. the fraction of trials containing no collisions.}
\label{fig:trees-collision-sr}
\end{figure}

We similarly present collision metrics in the generalization Trees environment in Figure \ref{fig:trees-collision-sr}; note that as this environment was not used for training, we do not present metrics for the expert policy. ViT models present the lowest collision rates, though ViT+LSTM is outperformed by ViT at 6m/s. We also notice the lack of a downward trend in the time spent in collision plot, perhaps an artifact of the trees being of uniform size and generally smaller than the spherical obstacles, causing the time spent in collision to be about a magnitude lower than in Spheres. Likely for similar reasons the success rates are higher, with ViT and ViT+LSTM outperforming other models, presenting 80\% success and 70\% success at 7m/s, respectively.

\begin{figure}
  \centering
  \begin{subfigure}{0.47\linewidth}
    \includegraphics[width=1.0\linewidth]{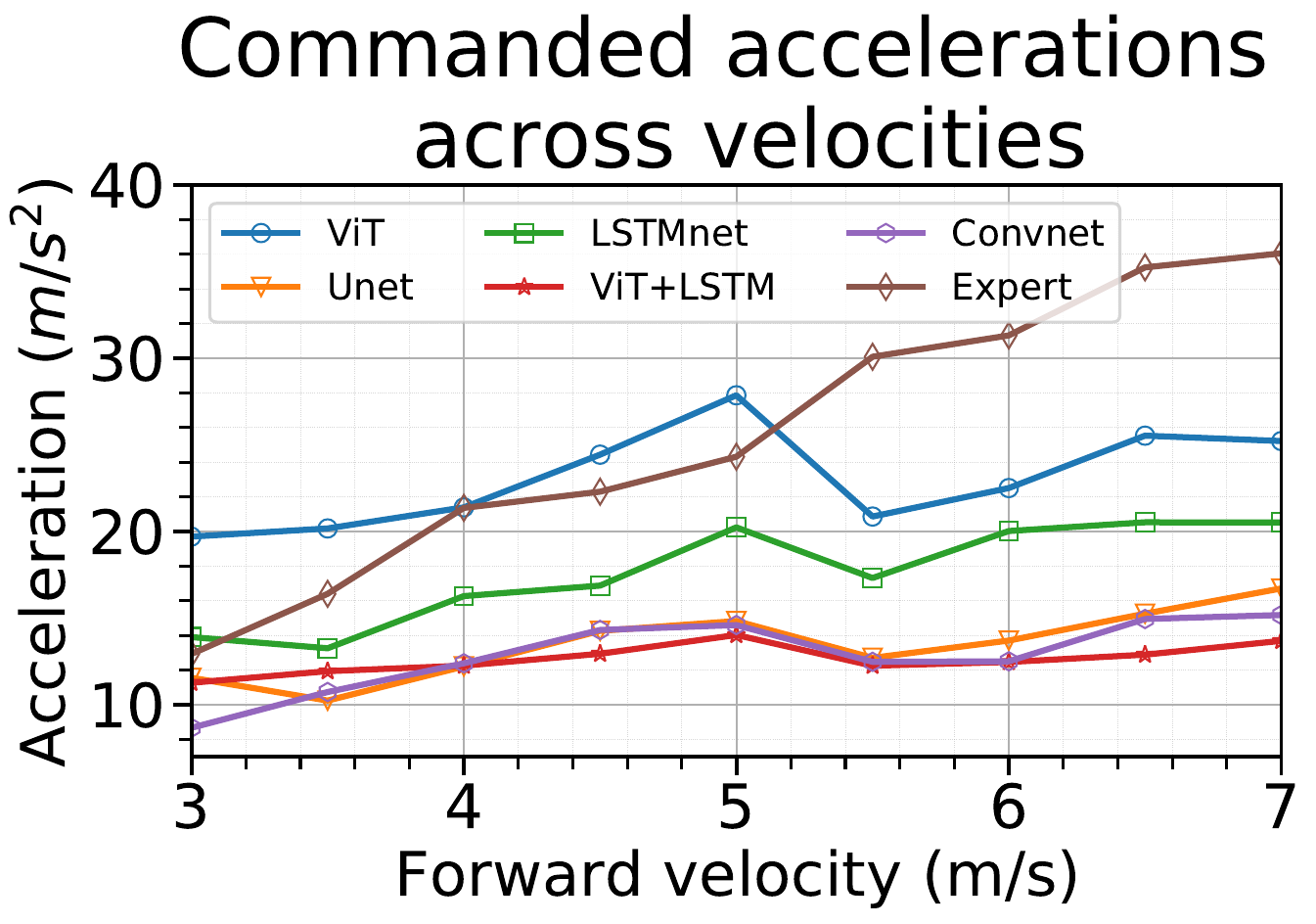}
    \vspace{-15pt}\caption{Spheres: drone acceleration.}
    \label{fig:spheres-acc}
  \end{subfigure}
  \hfill
  \begin{subfigure}{0.47\linewidth}
    \includegraphics[width=1.0\linewidth]{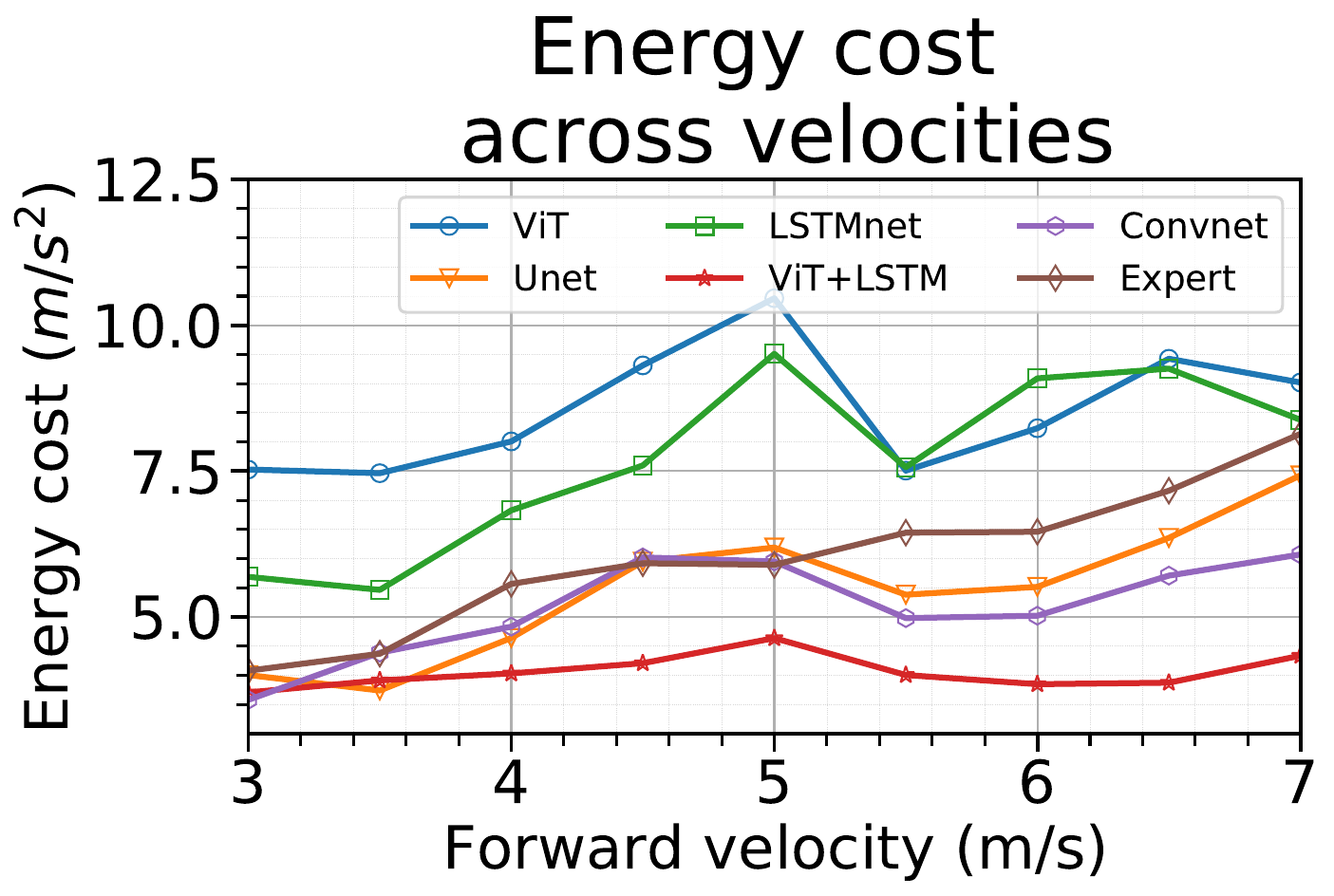}
    \vspace{-15pt}\caption{Spheres: energy cost.}
    \label{fig:spheres-energy}
  \end{subfigure}
  \hfill
  \begin{subfigure}{0.48\linewidth}
    \includegraphics[width=1.0\linewidth]{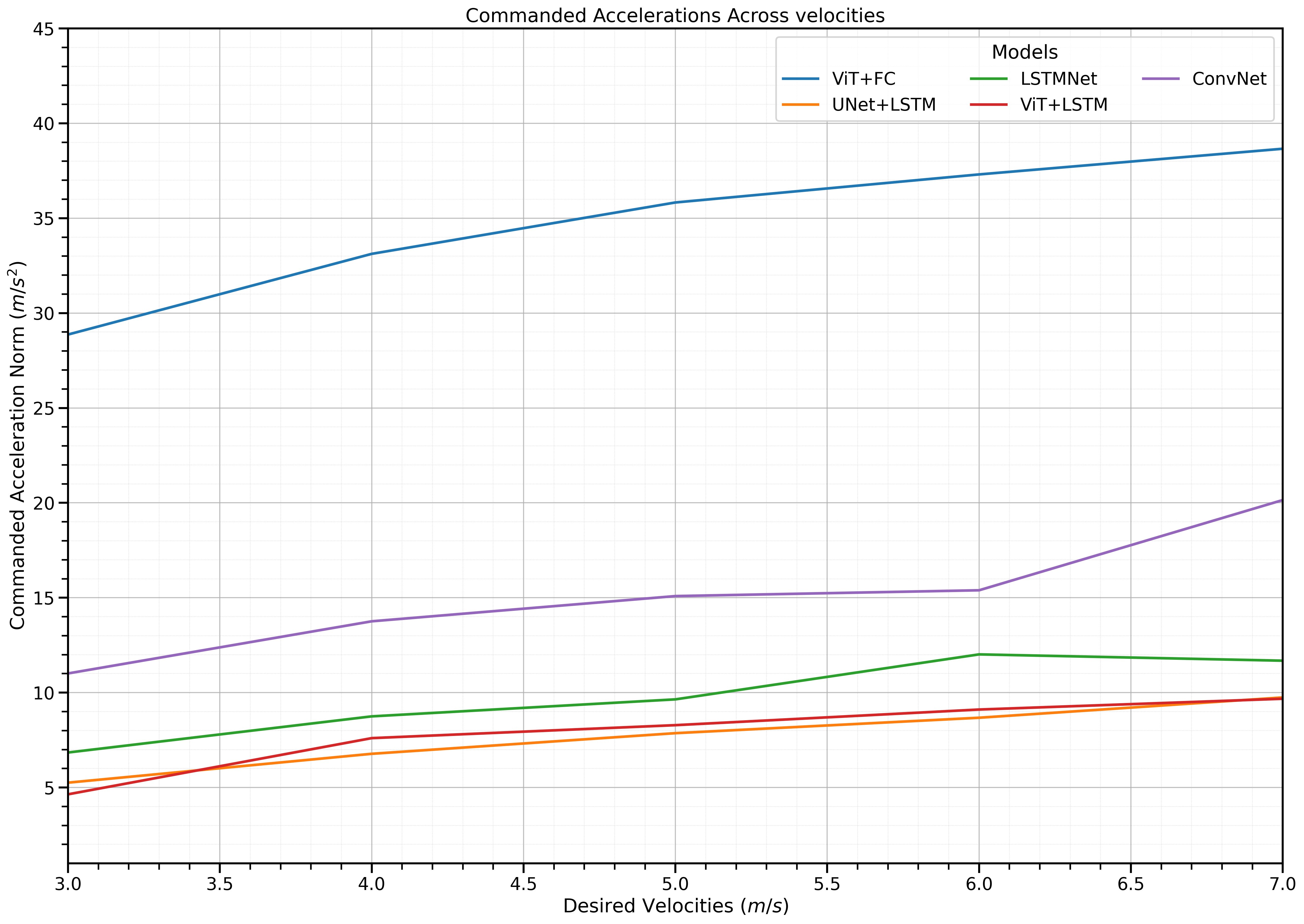}
    \vspace{-15pt}\caption{Trees: drone acceleration.}
    \label{fig:trees-acc}
  \end{subfigure}
  \hfill
  \begin{subfigure}{0.48\linewidth}
    \includegraphics[width=1.0\linewidth]{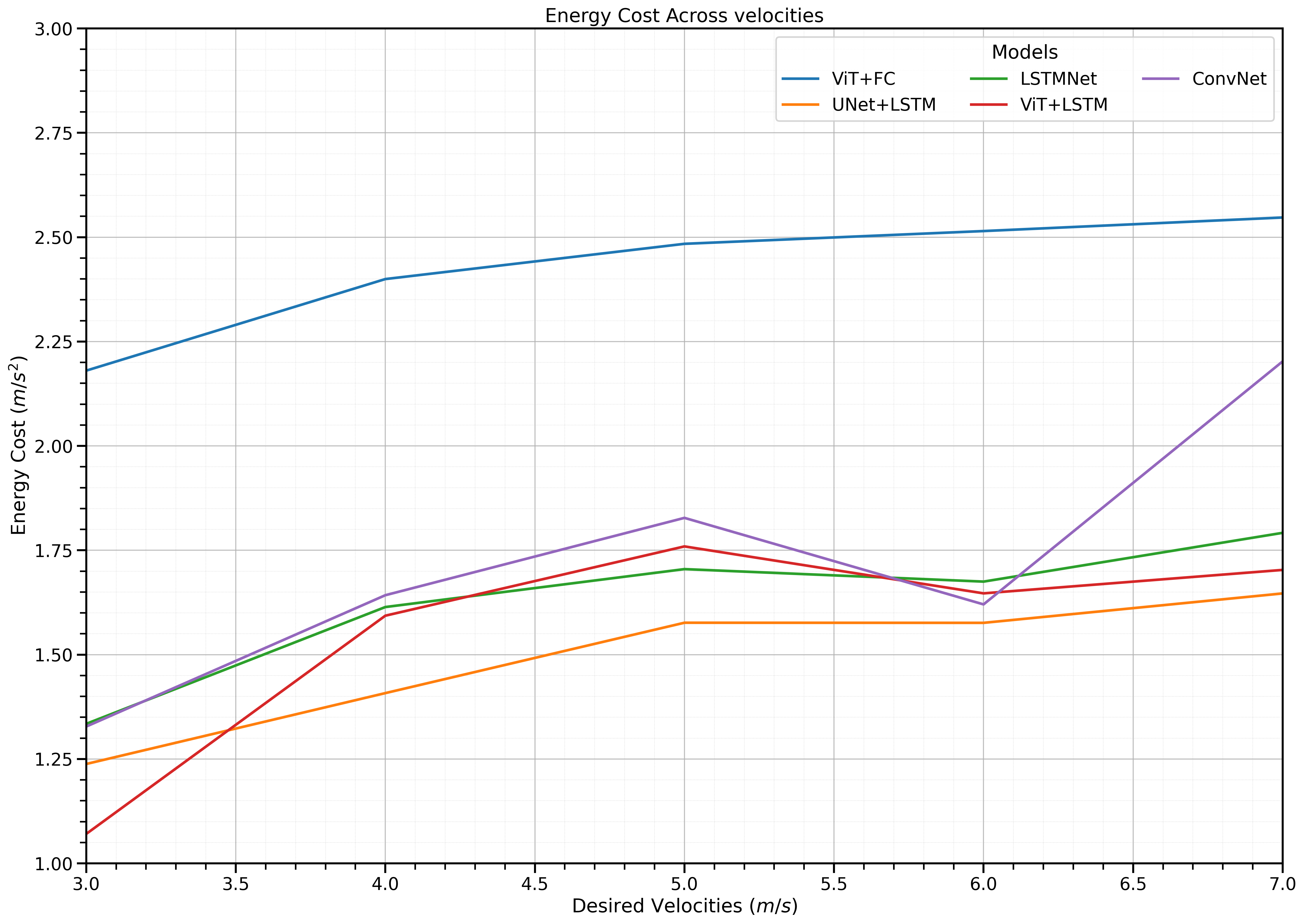}
    \vspace{-15pt}\caption{Trees: energy cost.}
    \label{fig:trees-energy}
    \end{subfigure}
    \caption{Commanded accelerations (\ref*{fig:spheres-acc}, \ref*{fig:trees-acc}) and computed energy costs (\ref*{fig:spheres-energy}, \ref*{fig:trees-energy}) for all models in the Spheres and Trees environments. Lower is considered better for both metrics. The ViT-only model commands high-acceleration values without achieving better performance, but the inclusion of recurrent layers in the ViT+LSTM model improves this behavior.}
    \label{fig:acc-energy-plots}
\end{figure}

Commanded accelerations and energy costs for Spheres and Trees environments are shown in Figure \ref{fig:acc-energy-plots}. In both environments, we see each metric improve as recurrence is added to the ViT model (becoming ViT+LSTM). The ViT+LSTM model in Spheres presents similar accelerations and energy costs as forward velocity increases, in contrast to the other models which generally increase with increasing forward speed. However, in the sparser Trees environment the models (barring ViT, also referred to as ViT+FC in these plots) have more similar characteristics.

%% file: tables/table-model-details.tex
\begin{table}
    \centering
    \begin{tabular}{lccccc}
        \hline  
        \multicolumn{1}{l}{\textbf{}} & \multicolumn{2}{c}{{\ul Structure}} & \multicolumn{3}{c}{{\ul Components}}\\
        \textbf{Model}    & \textbf{UNet} & \multicolumn{1}{c|}{\textbf{ViT}} & \textbf{Conv} & \textbf{LSTM} & \textbf{MLP} \\ \hline
        ConvNet        & \redcross        & \multicolumn{1}{c|}{\redcross}     &  \greencheck(2)      &  \redcross       & \greencheck (4)     \\ 
        LSTMnet           & \redcross        & \multicolumn{1}{c|}{\redcross}     &  \greencheck(2)      &  \greencheck(5)       & \greencheck (3)      \\ 
        UNet+LSTM         & \greencheck      & \multicolumn{1}{c|}{\redcross}     &  \greencheck(2)      &  \greencheck(2)       & \greencheck  (3)     \\ 
        ViT            & \redcross        & \multicolumn{1}{c|}{\greencheck}   &  \greencheck(1)        &  \redcross       & \greencheck (3)      \\ 
        ViT+LSTM          & \redcross        & \multicolumn{1}{c|}{\greencheck}   &  \greencheck(1)      &  \greencheck(3)       & \greencheck (2)      \\ \hline
    \end{tabular}
    \caption{Structure and layer components per model. ConvNet and LSTMnet are structured by the rudimentary layer types themselves. Numbers next to checkmarks indicate how many layers of that type are present in the model.}
    \label{tab:model-layers}
  \end{table}

%% file: tikz-nn-diagrams.tex
\begin{figure*}[!htb]
    \centering
    \begin{tikzpicture}
    
        \draw[green, fill=green!40] (-2.1, -0.25) -- (-2.1, 2.1) -- (-1.4, 2.6) -- (-1.4, 0.25) -- cycle;
        \draw[pink, fill=pink!60] (-1.8, 0.2) -- (-1.8, 1.8) -- (-1.2, 2.3) -- (-1.2, 0.65) -- cycle;
    
        \foreach \m/\l [count=\y] in {1,2,3,missing,4}
          \node [every neuron/.try, neuron \m/.try,] (input-\m) at (0,2.85 - 0.5 * \y) {};
        
        \foreach \m [count=\y] in {1,2,missing,3}
          \node [every neuron/.try, neuron \m/.try ] (hidden-\m) at (1,2.6-\y*0.5) {};
        
        \foreach \m [count=\y] in {1,2, 3}
          \node [every neuron/.try, neuron \m/.try ] (output-\m) at (2,2.35-0.5 * \y) {};
        
        \foreach \i in {1, 2, 3,4}
          \foreach \j in {1,2,3}
            \draw [->, very thin, color=brown] (input-\i) -- (hidden-\j);
        
        \foreach \i in {1,2,3}
          \foreach \j in {1,2,3}
            \draw [->, very thin, color=brown] (hidden-\i) -- (output-\j);

    \end{tikzpicture}
    \caption{ConvNet is composed of convolutional layers utilizing a combination of average and max pooling, and batched normalization, followed by fully connected layers.}
    \label{fig:tikz-convnet}
\end{figure*}

\begin{figure*}[!htb]
    \centering
    \begin{tikzpicture}
        \begin{scope}[scale=0.7, every node/.style={scale=0.7}]
        \draw[rounded corners, fill=yellow!30] (0, 0) rectangle (4.0, 2.5) {};
    
        \node[circle, fill=orange, minimum size=12pt] (fl-x) at (0.65, 2.1) {};
        \node at (0.65, 2.1) {$\times$};
    
        \draw[fill=orange] (0.45, 0.5) rectangle (0.85, 0.9);
        \node at (0.65, 0.7) {$\sigma$};
    
        \draw[fill=orange] (1.0, 0.5) rectangle (1.4, 0.9);
        \node at (1.2, 0.7) {$\sigma$};
    
        \node[circle, fill=orange, minimum size=12pt] (sl-+) at (1.9, 2.1) {};
        \node at (1.9, 2.1) {$+$};
    
        \node[circle, fill=orange, minimum size=12pt] (sl-x) at (1.9, 1.35) {};
        \node at (1.9, 1.35) {$\times$};
        
        \draw[fill=orange] (1.6, 0.5) rectangle (2.2, 0.9);
        \node at (1.9, 0.7) {\scriptsize$\tanh$};
    
        
        \draw[fill=orange] (2.4, 0.5) rectangle (2.8, 0.9);
        \node at (2.6, 0.7) {$\sigma$};
    
        \draw [black,fill=orange!20] (3.2,1.8) ellipse (8pt and 4pt);
        \node at (3.2, 1.8) {\tiny$\tanh$};
        
        \node[circle, fill=orange, minimum size=12pt] (fifl-x) at (3.2, 1.35) {};
        \node at (3.2, 1.35) {$\times$};
    
        \draw[-, very thick, rounded corners] (0.4, -0.2) |- (0.6, 0.25);
        \draw[-, very thick, rounded corners] (-0.25, 0.25) -- (2.6, 0.25) -| (2.6, 0.5);
        \draw[-, very thick,] (0.65, 0.25) -- (0.65, 0.5);
        \draw[-, very thick,] (1.2, 0.25) -- (1.2, 0.5);
        \draw[-, very thick,] (1.9, 0.25) -- (1.9, 0.5);
    
        \draw[->, very thick,] (0.65, 0.9) -- (fl-x);
    
        \draw[->, very thick, rounded corners] (1.2, 0.9) |- (sl-x);
    
        \draw[->, very thick,] (1.9, 0.9) -- (sl-x);
        \draw[->, very thick,] (sl-x) -- (sl-+);
    
        \draw[->, very thick, rounded corners] (2.6, 0.9) |- (fifl-x);
    
        \draw[->, very thick] (-0.25, 2.1) -- (fl-x) -- (sl-+) -- (4.3, 2.1);
    
        \draw[-, very thick] (3.2, 2.1) -- (3.2, 1.95);
        \draw[-, very thick] (3.2, 1.70) -- (fifl-x);
        \draw[->, very thick, rounded corners] (fifl-x) |- (4.3, 0.25);
        \draw[->, very thick, rounded corners] (3.3, 0.25) -| (3.7, 2.8);
    
        \begin{scope}[on background layer]
            \draw[rounded corners, fill=yellow!30] (0.1, 0.1) rectangle (4.1, 2.6) {};
        \end{scope}
        \end{scope}

        \tikzset{shift={(0.2,-0.4)}}  
        \draw[green, fill=green!40] (-2.1, -0.25) -- (-2.1, 2.1) -- (-1.4, 2.6) -- (-1.4, 0.25) -- cycle;
        \draw[pink, fill=pink!60] (-1.8, 0.2) -- (-1.8, 1.8) -- (-1.2, 2.3) -- (-1.2, 0.65) -- cycle;

        \tikzset{shift={(-0.9,0)}}  
        \foreach \m/\l [count=\y] in {1,2,3,missing,4}
          \node [every neuron/.try, neuron \m/.try,] (input-\m) at (5,2.85 - 0.5 * \y) {};
        
        \foreach \m [count=\y] in {1,2,missing,3}
          \node [every neuron/.try, neuron \m/.try ] (hidden-\m) at (6,2.6-\y*0.5) {};
        
        \foreach \m [count=\y] in {1,2, 3}
          \node [every neuron/.try, neuron \m/.try ] (output-\m) at (7,2.35-0.5 * \y) {};
        
        \foreach \i in {1, 2, 3,4}
          \foreach \j in {1,2,3}
            \draw [->, very thin, color=brown] (input-\i) -- (hidden-\j);
        
        \foreach \i in {1,2,3}
          \foreach \j in {1,2,3}
            \draw [->, very thin, color=brown] (hidden-\i) -- (output-\j);

    \end{tikzpicture}
    \caption{LSTMNet first uses convolutional layers (similar to Figure \ref{fig:tikz-convnet}), then a five-layer LSTM followed by fully-connected layers.}
    \label{fig:tikz-lstmnet}
\end{figure*}

\begin{figure*}[!htb]
    \centering
    \begin{tikzpicture}
        \tikzmath{\scaleEnc2=0.75;\shiftEnc2=0.2;
                  \scaleEnc3=0.5;\shiftEnc3=0.4;}
        \draw[-, very thin, fill=orange!40] (0, 0) -- (0, 1) -- (0.5, 1.25) -- (0.5, 0.25) -- cycle;
        \draw[-, very thin,] (0.05, 0) -- (0.05, 1) -- (0.55, 1.25) -- (0.55, 0.25) -- cycle;
        \draw[-, very thin, fill=orange!40] (0.1, 0) -- (0.1, 1) -- (0.6, 1.25) -- (0.6, 0.25) -- cycle;

        \draw[-, very thin,] (0,0) -- (0.05, 0) -- (0.1, 0);
        \draw[-, very thin,] (0,1) -- (0.05, 1) -- (0.1, 1);
        \draw[-, very thin,] (0.5,1.25) -- (0.55, 1.25) -- (0.6, 1.25);

        \draw[->, very thin,] (0.35, 0.625) -- (3.75, 0.625);
        
        \tikzset{shift={(0.5,-0.6)}};
        \draw[-, very thin, fill=orange!40] (0, 0) -- (0, 1*\scaleEnc2) -- (0.5*\scaleEnc2, 1.25*\scaleEnc2) -- (0.5*\scaleEnc2, 0.25*\scaleEnc2) -- cycle;
        \draw[-, very thin,] (0.05*\scaleEnc2+\shiftEnc2, 0) -- (0.05*\scaleEnc2+\shiftEnc2, 1*\scaleEnc2) -- (0.55*\scaleEnc2+\shiftEnc2, 1.25*\scaleEnc2) -- (0.55*\scaleEnc2+\shiftEnc2, 0.25*\scaleEnc2) -- cycle;
        \draw[-, very thin, fill=orange!40] (0.1*\scaleEnc2+2*\shiftEnc2, 0) -- (0.1*\scaleEnc2+2*\shiftEnc2, 1*\scaleEnc2) -- (0.6*\scaleEnc2+2*\shiftEnc2, 1.25*\scaleEnc2) -- (0.6*\scaleEnc2+2*\shiftEnc2, 0.25*\scaleEnc2) -- cycle;
        
        \draw[-, very thin,] (0,0) -- (0.05*\scaleEnc2+\shiftEnc2, 0) -- (0.1*\scaleEnc2+2*\shiftEnc2, 0);
        \draw[-, very thin,] (0,1*\scaleEnc2) -- (0.05*\scaleEnc2+\shiftEnc2, 1*\scaleEnc2) -- (0.1*\scaleEnc2+2*\shiftEnc2, 1*\scaleEnc2);
        \draw[-, very thin,] (0.5*\scaleEnc2,1.25*\scaleEnc2) -- (0.55*\scaleEnc2+\shiftEnc2, 1.25*\scaleEnc2) -- (0.6*\scaleEnc2+2*\shiftEnc2, 1.25*\scaleEnc2);

        \draw[-, very thin, fill=orange!40] (0,0) -- (0.05*\scaleEnc2+\shiftEnc2, 0) -- (0.05*\scaleEnc2+\shiftEnc2, 1*\scaleEnc2) -- (0,1*\scaleEnc2) -- cycle;
        \draw[-, very thin, fill=orange!40] (0.05*\scaleEnc2+\shiftEnc2, 0) -- (0.1*\scaleEnc2+2*\shiftEnc2, 0) -- (0.1*\scaleEnc2+2*\shiftEnc2, 1*\scaleEnc2) -- (0.05*\scaleEnc2+\shiftEnc2, 1*\scaleEnc2);
        \draw[-, very thin, fill=orange!40] (0,1*\scaleEnc2) -- (0.05*\scaleEnc2+\shiftEnc2, 1*\scaleEnc2) -- (0.55*\scaleEnc2+\shiftEnc2, 1.25*\scaleEnc2) -- (0.5*\scaleEnc2,1.25*\scaleEnc2) -- cycle;
        \draw[-, very thin, fill=orange!40] (0.05*\scaleEnc2+\shiftEnc2, 1*\scaleEnc2) -- (0.1*\scaleEnc2+2*\shiftEnc2, 1*\scaleEnc2) -- (0.6*\scaleEnc2+2*\shiftEnc2, 1.25*\scaleEnc2) -- (0.55*\scaleEnc2+\shiftEnc2, 1.25*\scaleEnc2) -- cycle;

        \draw[->, very thin,line width=0.01mm] (0.65, 0.5) -- (2.75, 0.5);

        \tikzset{shift={(0.7,-0.6)}};
        \draw[-, very thin, fill=orange!40] (0, 0) -- (0, 1*\scaleEnc3) -- (0.5*\scaleEnc3, 1.25*\scaleEnc3) -- (0.5*\scaleEnc3, 0.25*\scaleEnc3) -- cycle;
        \draw[-, very thin,] (0.05*\scaleEnc3+\shiftEnc3, 0) -- (0.05*\scaleEnc3+\shiftEnc3, 1*\scaleEnc3) -- (0.55*\scaleEnc3+\shiftEnc3, 1.25*\scaleEnc3) -- (0.55*\scaleEnc3+\shiftEnc3, 0.25*\scaleEnc3) -- cycle;
        \draw[-, very thin, fill=orange!40] (0.1*\scaleEnc3+2*\shiftEnc3, 0) -- (0.1*\scaleEnc3+2*\shiftEnc3, 1*\scaleEnc3) -- (0.6*\scaleEnc3+2*\shiftEnc3, 1.25*\scaleEnc3) -- (0.6*\scaleEnc3+2*\shiftEnc3, 0.25*\scaleEnc3) -- cycle;
        
        \draw[-, very thin,] (0,0) -- (0.05*\scaleEnc3+\shiftEnc3, 0) -- (0.1*\scaleEnc3+2*\shiftEnc3, 0);
        \draw[-, very thin,] (0,1*\scaleEnc3) -- (0.05*\scaleEnc3+\shiftEnc3, 1*\scaleEnc3) -- (0.1*\scaleEnc3+2*\shiftEnc3, 1*\scaleEnc3);
        \draw[-, very thin,] (0.5*\scaleEnc3,1.25*\scaleEnc3) -- (0.55*\scaleEnc3+\shiftEnc3, 1.25*\scaleEnc3) -- (0.6*\scaleEnc3+2*\shiftEnc3, 1.25*\scaleEnc3);

        \draw[-, very thin, fill=orange!40] (0,0) -- (0.05*\scaleEnc3+\shiftEnc3, 0) -- (0.05*\scaleEnc3+\shiftEnc3, 1*\scaleEnc3) -- (0,1*\scaleEnc3) -- cycle;
        \draw[-, very thin, fill=orange!40] (0.05*\scaleEnc3+\shiftEnc3, 0) -- (0.1*\scaleEnc3+2*\shiftEnc3, 0) -- (0.1*\scaleEnc3+2*\shiftEnc3, 1*\scaleEnc3) -- (0.05*\scaleEnc3+\shiftEnc3, 1*\scaleEnc3);
        \draw[-, very thin, fill=orange!40] (0,1*\scaleEnc3) -- (0.05*\scaleEnc3+\shiftEnc3, 1*\scaleEnc3) -- (0.55*\scaleEnc3+\shiftEnc3, 1.25*\scaleEnc3) -- (0.5*\scaleEnc3,1.25*\scaleEnc3) -- cycle;
        \draw[-, very thin, fill=orange!40] (0.05*\scaleEnc3+\shiftEnc3, 1*\scaleEnc3) -- (0.1*\scaleEnc3+2*\shiftEnc3, 1*\scaleEnc3) -- (0.6*\scaleEnc3+2*\shiftEnc3, 1.25*\scaleEnc3) -- (0.55*\scaleEnc3+\shiftEnc3, 1.25*\scaleEnc3) -- cycle;

        \tikzset{shift={(1.3,0.6)}};
        \draw[-, very thin, fill=orange!40] (0, 0) -- (0, 1*\scaleEnc2) -- (0.5*\scaleEnc2, 1.25*\scaleEnc2) -- (0.5*\scaleEnc2, 0.25*\scaleEnc2) -- cycle;
        \draw[-, very thin,] (0.05*\scaleEnc2+\shiftEnc2, 0) -- (0.05*\scaleEnc2+\shiftEnc2, 1*\scaleEnc2) -- (0.55*\scaleEnc2+\shiftEnc2, 1.25*\scaleEnc2) -- (0.55*\scaleEnc2+\shiftEnc2, 0.25*\scaleEnc2) -- cycle;
        \draw[-, very thin, fill=orange!40] (0.1*\scaleEnc2+2*\shiftEnc2, 0) -- (0.1*\scaleEnc2+2*\shiftEnc2, 1*\scaleEnc2) -- (0.6*\scaleEnc2+2*\shiftEnc2, 1.25*\scaleEnc2) -- (0.6*\scaleEnc2+2*\shiftEnc2, 0.25*\scaleEnc2) -- cycle;
        
        \draw[-, very thin,] (0,0) -- (0.05*\scaleEnc2+\shiftEnc2, 0) -- (0.1*\scaleEnc2+2*\shiftEnc2, 0);
        \draw[-, very thin,] (0,1*\scaleEnc2) -- (0.05*\scaleEnc2+\shiftEnc2, 1*\scaleEnc2) -- (0.1*\scaleEnc2+2*\shiftEnc2, 1*\scaleEnc2);
        \draw[-, very thin,] (0.5*\scaleEnc2,1.25*\scaleEnc2) -- (0.55*\scaleEnc2+\shiftEnc2, 1.25*\scaleEnc2) -- (0.6*\scaleEnc2+2*\shiftEnc2, 1.25*\scaleEnc2);

        \draw[-, very thin, fill=orange!40] (0,0) -- (0.05*\scaleEnc2+\shiftEnc2, 0) -- (0.05*\scaleEnc2+\shiftEnc2, 1*\scaleEnc2) -- (0,1*\scaleEnc2) -- cycle;
        \draw[-, very thin, fill=orange!40] (0.05*\scaleEnc2+\shiftEnc2, 0) -- (0.1*\scaleEnc2+2*\shiftEnc2, 0) -- (0.1*\scaleEnc2+2*\shiftEnc2, 1*\scaleEnc2) -- (0.05*\scaleEnc2+\shiftEnc2, 1*\scaleEnc2);
        \draw[-, very thin, fill=orange!40] (0,1*\scaleEnc2) -- (0.05*\scaleEnc2+\shiftEnc2, 1*\scaleEnc2) -- (0.55*\scaleEnc2+\shiftEnc2, 1.25*\scaleEnc2) -- (0.5*\scaleEnc2,1.25*\scaleEnc2) -- cycle;
        \draw[-, very thin, fill=orange!40] (0.05*\scaleEnc2+\shiftEnc2, 1*\scaleEnc2) -- (0.1*\scaleEnc2+2*\shiftEnc2, 1*\scaleEnc2) -- (0.6*\scaleEnc2+2*\shiftEnc2, 1.25*\scaleEnc2) -- (0.55*\scaleEnc2+\shiftEnc2, 1.25*\scaleEnc2) -- cycle;

        \tikzset{shift={(1.0,0.5)}};
        \draw[-, very thin, fill=orange!40] (0, 0) -- (0, 1) -- (0.5, 1.25) -- (0.5, 0.25) -- cycle;
        \draw[-, very thin,] (0.05, 0) -- (0.05, 1) -- (0.55, 1.25) -- (0.55, 0.25) -- cycle;
        \draw[-, very thin, fill=orange!40] (0.1, 0) -- (0.1, 1) -- (0.6, 1.25) -- (0.6, 0.25) -- cycle;

        \draw[-, very thin,] (0,0) -- (0.05, 0) -- (0.1, 0);
        \draw[-, very thin,] (0,1) -- (0.05, 1) -- (0.1, 1);
        \draw[-, very thin,] (0.5,1.25) -- (0.55, 1.25) -- (0.6, 1.25);

        \tikzset{shift={(4.0,-1.0)}};
        \draw[green, fill=green!40] (-2.1, -0.25) -- (-2.1, 2.1) -- (-1.4, 2.6) -- (-1.4, 0.25) -- cycle;
        \draw[pink, fill=pink!60] (-1.8, 0.2) -- (-1.8, 1.8) -- (-1.2, 2.3) -- (-1.2, 0.65) -- cycle;

        \tikzset{shift={(0, 0.3)}}
        \begin{scope}[scale=0.7, every node/.style={scale=0.7}]
            \draw[rounded corners, fill=yellow!30] (0, 0) rectangle (4.0, 2.5) {};
        
            \node[circle, fill=orange, minimum size=12pt] (fl-x) at (0.65, 2.1) {};
            \node at (0.65, 2.1) {$\times$};
        
            \draw[fill=orange] (0.45, 0.5) rectangle (0.85, 0.9);
            \node at (0.65, 0.7) {$\sigma$};
        
            \draw[fill=orange] (1.0, 0.5) rectangle (1.4, 0.9);
            \node at (1.2, 0.7) {$\sigma$};
        
            \node[circle, fill=orange, minimum size=12pt] (sl-+) at (1.9, 2.1) {};
            \node at (1.9, 2.1) {$+$};
        
            \node[circle, fill=orange, minimum size=12pt] (sl-x) at (1.9, 1.35) {};
            \node at (1.9, 1.35) {$\times$};
            
            \draw[fill=orange] (1.6, 0.5) rectangle (2.2, 0.9);
            \node at (1.9, 0.7) {\scriptsize$\tanh$};
        
            
            \draw[fill=orange] (2.4, 0.5) rectangle (2.8, 0.9);
            \node at (2.6, 0.7) {$\sigma$};
        
            \draw [black,fill=orange!20] (3.2,1.8) ellipse (8pt and 4pt);
            \node at (3.2, 1.8) {\tiny$\tanh$};
            
            \node[circle, fill=orange, minimum size=12pt] (fifl-x) at (3.2, 1.35) {};
            \node at (3.2, 1.35) {$\times$};
        
            \draw[-, very thick, rounded corners] (0.4, -0.2) |- (0.6, 0.25);
            \draw[-, very thick, rounded corners] (-0.25, 0.25) -- (2.6, 0.25) -| (2.6, 0.5);
            \draw[-, very thick,] (0.65, 0.25) -- (0.65, 0.5);
            \draw[-, very thick,] (1.2, 0.25) -- (1.2, 0.5);
            \draw[-, very thick,] (1.9, 0.25) -- (1.9, 0.5);
        
            \draw[->, very thick,] (0.65, 0.9) -- (fl-x);
        
            \draw[->, very thick, rounded corners] (1.2, 0.9) |- (sl-x);
        
            \draw[->, very thick,] (1.9, 0.9) -- (sl-x);
            \draw[->, very thick,] (sl-x) -- (sl-+);
        
            \draw[->, very thick, rounded corners] (2.6, 0.9) |- (fifl-x);
        
            \draw[->, very thick] (-0.25, 2.1) -- (fl-x) -- (sl-+) -- (4.3, 2.1);
        
            \draw[-, very thick] (3.2, 2.1) -- (3.2, 1.95);
            \draw[-, very thick] (3.2, 1.70) -- (fifl-x);
            \draw[->, very thick, rounded corners] (fifl-x) |- (4.3, 0.25);
            \draw[->, very thick, rounded corners] (3.3, 0.25) -| (3.7, 2.8);
        
            \begin{scope}[on background layer]
                \draw[rounded corners, fill=yellow!30] (0.1, 0.1) rectangle (4.1, 2.6) {};
            \end{scope}
        \end{scope}

        \tikzset{shift={(-0.65,-0.5)}}        
        \foreach \m/\l [count=\y] in {1,2,3,missing,4}
          \node [every neuron/.try, neuron \m/.try,] (input-\m) at (5,2.85 - 0.5 * \y) {};
        
        \foreach \m [count=\y] in {1,2,missing,3}
          \node [every neuron/.try, neuron \m/.try ] (hidden-\m) at (6,2.6-\y*0.5) {};
        
        \foreach \m [count=\y] in {1,2, 3}
          \node [every neuron/.try, neuron \m/.try ] (output-\m) at (7,2.35-0.5 * \y) {};
        
        \foreach \i in {1, 2, 3,4}
          \foreach \j in {1,2,3}
            \draw [->, very thin, color=brown] (input-\i) -- (hidden-\j);
        
        \foreach \i in {1,2,3}
          \foreach \j in {1,2,3}
            \draw [->, very thin, color=brown] (hidden-\i) -- (output-\j);
    \end{tikzpicture}
    \caption{UNet+LSTM is composed of a three-layer-deep UNet (2D convolutions and transposed convolutions with max pooling) but with skip connections that match each decoder-stage dimensionality, avoiding cropping or interpolating encoding-stage signals, followed by a two-layer LSTMNet (Figure \ref{fig:tikz-lstmnet}).}
    \label{fig:tikz-unetlstm}
\end{figure*}

\input{tikz-vit}

\input{tikz-vitlstm}

%% file: tikz-vit.tex
\begin{figure*}[!htb]
    \centering
    \resizebox{1.2\columnwidth}{!}{%
    \begin{tikzpicture}
        \begin{scope}
        \begin{tikzpicture}[scale=0.6, every node/.style={scale=0.6}, shift={(2, 0)}]
            \coordinate (input) at (-1.0, 0);
            \node at (-1.0, 0.5) {Depth Image};
            \draw[->] (-1.0, 0.3) -- (-1.0, -0.5);
            \draw[rounded corners, fill=gray!30] (-2.5, -4.5) rectangle (0.5, -0.5);
            \node [fill=yellow!40, draw, minimum size=0.5cm, rotate=0, rounded corners, below of=input] (block1) {Patch Embedding};
            \node [fill=yellow!40, draw, minimum size=0.5cm, rotate=0, rounded corners, below of=block1] (block2) {Efficient Self-Attn};
            \node [fill=yellow!40, draw, minimum size=0.5cm, rotate=0, rounded corners, below of=block2] (block3) {Mix FFN};
            \node [fill=yellow!40, draw, minimum size=0.5cm, rotate=0, rounded corners, below of=block3] (block4) {Patch Merging};

            \draw[->] (block1) -- (block2);
            \draw[->] (block2) -- (block3);
            \draw[->] (block3) -- (block4);
            
            \draw[red, fill=red!40,] (1.1, -3.4) rectangle (3.7, -1.6);
            \node at (2.4, -3.7) {Encoding};
            \tikzset{shift={(7.0, 0.)}}
            \coordinate (input) at (-1.0, 0);
            \draw[rounded corners, fill=gray!30] (-2.5, -4.5) rectangle (0.5, -0.5);
            \node [fill=yellow!40, draw, minimum size=0.5cm, rotate=0, rounded corners, below of=input] (block1) {Patch Embedding};
            \node [fill=yellow!40, draw, minimum size=0.5cm, rotate=0, rounded corners, below of=block1] (block2) {Efficient Self-Attn};
            \node [fill=yellow!40, draw, minimum size=0.5cm, rotate=0, rounded corners, below of=block2] (block3) {Mix FFN};
            \node [fill=yellow!40, draw, minimum size=0.5cm, rotate=0, rounded corners, below of=block3] (block4) {Patch Merging};

            \draw[->] (block1) -- (block2);
            \draw[->] (block2) -- (block3);
            \draw[->] (block3) -- (block4);
        \end{tikzpicture}
        \end{scope}

        \draw[-<, dashed, thick] (3.0, 1.9) |- (7.3, 3.0) |- (7.3, 2.4);

        \tikzset{shift={(2, -0.25)}}
        \foreach \y in {0.1,0.2,...,.9} {
          \foreach \x in {0.1,0.2,...,1.5} {
              \pgfmathparse{0.9*rnd+0.3}
              \definecolor{MyColor}{rgb}{\pgfmathresult,0.,0.}
              \node[fill=MyColor!60,inner sep=0.1cm,outer sep=0pt,anchor=center] at (4.5 + \x,1 + \y) {}; 
          }
        }

        \node[circle, fill=orange, minimum size=10pt] (sl-+) at (5.3, 2.3) {};
        \node at (5.3, 2.3) {$+$};
        \node at (5.3, 0.75) {\scriptsize Pixel Shuffle};

        \tikzset{shift={(7,0.65)}}

        \tikzset{shift={(-5.4,-0.5)}}        
        \foreach \m/\l [count=\y] in {1,2,3,missing,4}
          \node [every neuron/.try, neuron \m/.try,] (input-\m) at (5,2.85 - 0.5 * \y) {};
        
        \foreach \m [count=\y] in {1,2,missing,3}
          \node [every neuron/.try, neuron \m/.try ] (hidden-\m) at (6,2.6-\y*0.5) {};
        
        \foreach \m [count=\y] in {1,2, 3}
          \node [every neuron/.try, neuron \m/.try ] (output-\m) at (7,2.35-0.5 * \y) {};
        
        \foreach \i in {1, 2, 3,4}
          \foreach \j in {1,2,3}
            \draw [->, very thin, color=brown] (input-\i) -- (hidden-\j);
        
        \foreach \i in {1,2,3}
          \foreach \j in {1,2,3}
            \draw [->, very thin, color=brown] (hidden-\i) -- (output-\j);
    \end{tikzpicture}%
    }
    \vspace{-2.cm}
    \caption{ViT is based on Segformer as described in the main text.}
    \label{fig:tikz-vit}
\end{figure*}

%% file: tikz-vitlstm.tex
\begin{figure*}[!htb]
    \centering
    \resizebox{2.1\columnwidth}{!}{%
    \begin{tikzpicture}
        \begin{scope}
        \begin{tikzpicture}[scale=0.6, every node/.style={scale=0.6}, shift={(2, 0)}]
            \coordinate (input) at (-1.0, 0);
            \node at (-1.0, 0.5) {Depth Image};
            \draw[->] (-1.0, 0.3) -- (-1.0, -0.5);
            \draw[rounded corners, fill=gray!30] (-2.5, -4.5) rectangle (0.5, -0.5);
            \node [fill=yellow!40, draw, minimum size=0.5cm, rotate=0, rounded corners, below of=input] (block1) {Patch Embedding};
            \node [fill=yellow!40, draw, minimum size=0.5cm, rotate=0, rounded corners, below of=block1] (block2) {Efficient Self-Attn};
            \node [fill=yellow!40, draw, minimum size=0.5cm, rotate=0, rounded corners, below of=block2] (block3) {Mix FFN};
            \node [fill=yellow!40, draw, minimum size=0.5cm, rotate=0, rounded corners, below of=block3] (block4) {Patch Merging};

            \draw[->] (block1) -- (block2);
            \draw[->] (block2) -- (block3);
            \draw[->] (block3) -- (block4);
            
            \draw[red, fill=red!40,] (1.1, -3.4) rectangle (3.7, -1.6);
            \node at (2.4, -3.7) {Encoding};
            \tikzset{shift={(7.0, 0.)}}
            \coordinate (input) at (-1.0, 0);
            \draw[rounded corners, fill=gray!30] (-2.5, -4.5) rectangle (0.5, -0.5);
            \node [fill=yellow!40, draw, minimum size=0.5cm, rotate=0, rounded corners, below of=input] (block1) {Patch Embedding};
            \node [fill=yellow!40, draw, minimum size=0.5cm, rotate=0, rounded corners, below of=block1] (block2) {Efficient Self-Attn};
            \node [fill=yellow!40, draw, minimum size=0.5cm, rotate=0, rounded corners, below of=block2] (block3) {Mix FFN};
            \node [fill=yellow!40, draw, minimum size=0.5cm, rotate=0, rounded corners, below of=block3] (block4) {Patch Merging};

            \draw[->] (block1) -- (block2);
            \draw[->] (block2) -- (block3);
            \draw[->] (block3) -- (block4);
        \end{tikzpicture}
        \end{scope}

        \draw[-<, dashed, thick] (3.0, 1.9) |- (7.3, 3.0) |- (7.3, 2.4);

        \tikzset{shift={(2, -0.25)}}
        \foreach \y in {0.1,0.2,...,.9} {
          \foreach \x in {0.1,0.2,...,1.5} {
              \pgfmathparse{0.9*rnd+0.3}
              \definecolor{MyColor}{rgb}{\pgfmathresult,0.,0.}
              \node[fill=MyColor!60,inner sep=0.1cm,outer sep=0pt,anchor=center] at (4.5 + \x,1 + \y) {}; 
          }
        }

        \node[circle, fill=orange, minimum size=10pt] (sl-+) at (5.3, 2.3) {};
        \node at (5.3, 2.3) {$+$};
        \node at (5.3, 0.75) {\scriptsize Pixel Shuffle};

        \tikzset{shift={(7,0.65)}}

        \tikzset{shift={(-5.4,-0.5)}}        
        \foreach \m/\l [count=\y] in {1,2,3,missing,4}
          \node [every neuron/.try, neuron \m/.try,] (input-\m) at (5,2.85 - 0.5 * \y) {};
        
        \foreach \m [count=\y] in {1,2,missing,3}
          \node [every neuron/.try, neuron \m/.try ] (hidden-\m) at (6,2.6-\y*0.5) {};
        
        \foreach \m [count=\y] in {1,2, 3}
          \node [every neuron/.try, neuron \m/.try ] (output-\m) at (7,2.35-0.5 * \y) {};
        
        \foreach \i in {1, 2, 3,4}
          \foreach \j in {1,2,3}
            \draw [->, very thin, color=brown] (input-\i) -- (hidden-\j);
        
        \foreach \i in {1,2,3}
          \foreach \j in {1,2,3}
            \draw [->, very thin, color=brown] (hidden-\i) -- (output-\j);

        \tikzset{shift={(10.0,0.0)}};
        \draw[green, fill=green!40] (-2.1, -0.25) -- (-2.1, 2.1) -- (-1.4, 2.6) -- (-1.4, 0.25) -- cycle;
        \draw[pink, fill=pink!60] (-1.8, 0.2) -- (-1.8, 1.8) -- (-1.2, 2.3) -- (-1.2, 0.65) -- cycle;

        \tikzset{shift={(-0.2, 0.3)}}
        \begin{scope}[scale=0.7, every node/.style={scale=0.7}]
            \draw[rounded corners, fill=yellow!30] (0, 0) rectangle (4.0, 2.5) {};
        
            \node[circle, fill=orange, minimum size=12pt] (fl-x) at (0.65, 2.1) {};
            \node at (0.65, 2.1) {$\times$};
        
            \draw[fill=orange] (0.45, 0.5) rectangle (0.85, 0.9);
            \node at (0.65, 0.7) {$\sigma$};
        
            \draw[fill=orange] (1.0, 0.5) rectangle (1.4, 0.9);
            \node at (1.2, 0.7) {$\sigma$};
        
            \node[circle, fill=orange, minimum size=12pt] (sl-+) at (1.9, 2.1) {};
            \node at (1.9, 2.1) {$+$};
        
            \node[circle, fill=orange, minimum size=12pt] (sl-x) at (1.9, 1.35) {};
            \node at (1.9, 1.35) {$\times$};
            
            \draw[fill=orange] (1.6, 0.5) rectangle (2.2, 0.9);
            \node at (1.9, 0.7) {\scriptsize$\tanh$};
        
            
            \draw[fill=orange] (2.4, 0.5) rectangle (2.8, 0.9);
            \node at (2.6, 0.7) {$\sigma$};
        
            \draw [black,fill=orange!20] (3.2,1.8) ellipse (8pt and 4pt);
            \node at (3.2, 1.8) {\tiny$\tanh$};
            
            \node[circle, fill=orange, minimum size=12pt] (fifl-x) at (3.2, 1.35) {};
            \node at (3.2, 1.35) {$\times$};
        
            \draw[-, very thick, rounded corners] (0.4, -0.2) |- (0.6, 0.25);
            \draw[-, very thick, rounded corners] (-0.25, 0.25) -- (2.6, 0.25) -| (2.6, 0.5);
            \draw[-, very thick,] (0.65, 0.25) -- (0.65, 0.5);
            \draw[-, very thick,] (1.2, 0.25) -- (1.2, 0.5);
            \draw[-, very thick,] (1.9, 0.25) -- (1.9, 0.5);
        
            \draw[->, very thick,] (0.65, 0.9) -- (fl-x);
        
            \draw[->, very thick, rounded corners] (1.2, 0.9) |- (sl-x);
        
            \draw[->, very thick,] (1.9, 0.9) -- (sl-x);
            \draw[->, very thick,] (sl-x) -- (sl-+);
        
            \draw[->, very thick, rounded corners] (2.6, 0.9) |- (fifl-x);
        
            \draw[->, very thick] (-0.25, 2.1) -- (fl-x) -- (sl-+) -- (4.3, 2.1);
        
            \draw[-, very thick] (3.2, 2.1) -- (3.2, 1.95);
            \draw[-, very thick] (3.2, 1.70) -- (fifl-x);
            \draw[->, very thick, rounded corners] (fifl-x) |- (4.3, 0.25);
            \draw[->, very thick, rounded corners] (3.3, 0.25) -| (3.7, 2.8);
        
            \begin{scope}[on background layer]
                \draw[rounded corners, fill=yellow!30] (0.1, 0.1) rectangle (4.1, 2.6) {};
            \end{scope}
        \end{scope}

        \tikzset{shift={(-1.0,-0.5)}}        
        \foreach \m/\l [count=\y] in {1,2,3,missing,4}
          \node [every neuron/.try, neuron \m/.try,] (input-\m) at (5,2.85 - 0.5 * \y) {};
        
        \foreach \m [count=\y] in {1,2,missing,3}
          \node [every neuron/.try, neuron \m/.try ] (hidden-\m) at (6,2.6-\y*0.5) {};
        
        \foreach \m [count=\y] in {1,2, 3}
          \node [every neuron/.try, neuron \m/.try ] (output-\m) at (7,2.35-0.5 * \y) {};
        
        \foreach \i in {1, 2, 3,4}
          \foreach \j in {1,2,3}
            \draw [->, very thin, color=brown] (input-\i) -- (hidden-\j);
        
        \foreach \i in {1,2,3}
          \foreach \j in {1,2,3}
            \draw [->, very thin, color=brown] (hidden-\i) -- (output-\j);
    \end{tikzpicture}%
    }
    \vspace{-2.5cm}
    \caption{ViT+LSTM utilizes a three-layer LSTMNet (Figure \ref{fig:tikz-lstmnet}) after the ViT (Figure \ref{fig:tikz-vit}).}
    \label{fig:tikz-vitlstm}
\end{figure*}

%% file: bib.bib
@inproceedings{segformer,
  title={SegFormer: Simple and Efficient Design for Semantic Segmentation with Transformers},
  author={Xie, Enze and Wang, Wenhai and Yu, Zhiding and Anandkumar, Anima and Alvarez, Jose M and Luo, Ping},
  booktitle={Neural Information Processing Systems (NeurIPS)},
  year={2021}
}

@software{torchbench,
author = {Constable, Will and Zhao, Xu and Bittorf, Victor and Christoffersen, Eric and Robie, Taylor and Han, Eric and Wu, Peng and Korovaiko, Nick and Ansel, Jason and Reblitz-Richardson, Orion and Chintala, Soumith},
month = sep,
title = {{TorchBench: A collection of open source benchmarks for PyTorch performance and usability evaluation}},
url = {https://github.com/pytorch/benchmark},
year = {2020}
}

@INPROCEEDINGS{pxshuffle,
  author={Shi, Wenzhe and Caballero, Jose and Huszár, Ferenc and Totz, Johannes and Aitken, Andrew P. and Bishop, Rob and Rueckert, Daniel and Wang, Zehan},
  booktitle={2016 IEEE Conference on Computer Vision and Pattern Recognition (CVPR)}, 
  title={Real-Time Single Image and Video Super-Resolution Using an Efficient Sub-Pixel Convolutional Neural Network}, 
  year={2016},
  volume={},
  number={},
  pages={1874-1883},
  doi={10.1109/CVPR.2016.207}}

@article{lstm,
  title={Long short-term memory},
  author={Hochreiter, Sepp and Schmidhuber, J{\"u}rgen},
  journal={Neural computation},
  volume={9},
  number={8},
  pages={1735--1780},
  year={1997},
  publisher={MIT press}
}

@article{transformers,
  title={Attention is all you need},
  author={Vaswani, Ashish and Shazeer, Noam and Parmar, Niki and Uszkoreit, Jakob and Jones, Llion and Gomez, Aidan N and Kaiser, {\L}ukasz and Polosukhin, Illia},
  journal={Advances in neural information processing systems},
  volume={30},
  year={2017}
}

@article{attention,
  title={Neural machine translation by jointly learning to align and translate},
  author={Bahdanau, Dzmitry and Cho, Kyunghyun and Bengio, Yoshua},
  journal={arXiv preprint arXiv:1409.0473},
  year={2014}
}

@article{vits,
  title={An image is worth 16x16 words: Transformers for image recognition at scale},
  author={Dosovitskiy, Alexey and Beyer, Lucas and Kolesnikov, Alexander and Weissenborn, Dirk and Zhai, Xiaohua and Unterthiner, Thomas and Dehghani, Mostafa and Minderer, Matthias and Heigold, Georg and Gelly, Sylvain and others},
  journal={arXiv preprint arXiv:2010.11929},
  year={2020}
}

@inproceedings{flightmare,
    title={Flightmare: A Flexible Quadrotor Simulator},
    author={Song, Yunlong and Naji, Selim and Kaufmann, Elia and Loquercio, Antonio and Scaramuzza, Davide},
    booktitle={Conference on Robot Learning},
    year={2020}
}

@online{dodgedrone-competition,
    title = {DodgeDrone: Vision-based Agile Drone Flight (ICRA 2022 Competition)},
    author = {Song, Yunlong and Kaufmann, Elia and Bauersfeld, Leonard and Loquercio, Antonio and Scaramuzza, Davide},
    year = {2022},
    url = {https://uzh-rpg.github.io/icra2022-dodgedrone/},
    note = {[Accessed on 12-02-2024]}
}

@inproceedings{convnet-housedigits,
  title={Convolutional neural networks applied to house numbers digit classification},
  author={Sermanet, Pierre and Chintala, Soumith and LeCun, Yann},
  booktitle={Proceedings of the 21st international conference on pattern recognition (ICPR2012)},
  pages={3288--3291},
  year={2012},
  organization={IEEE}
}

@article{convnet-original-lecun,
  title={Handwritten digit recognition with a back-propagation network},
  author={LeCun, Yann and Boser, Bernhard and Denker, John and Henderson, Donnie and Howard, Richard and Hubbard, Wayne and Jackel, Lawrence},
  journal={Advances in neural information processing systems},
  volume={2},
  year={1989}
}

@article{ye2023real,
  title={Real-Time Object Detection Network in UAV-Vision Based on CNN and Transformer},
  author={Ye, Tao and Qin, Wenyang and Zhao, Zongyang and Gao, Xiaozhi and Deng, Xiangpeng and Ouyang, Yu},
  journal={IEEE Transactions on Instrumentation and Measurement},
  volume={72},
  pages={1--13},
  year={2023},
  publisher={IEEE}
}

@article{sun2022siamese,
  title={Siamese Transformer Network: Building an autonomous real-time target tracking system for UAV},
  author={Sun, Xiaolou and Wang, Qi and Xie, Fei and Quan, Zhibin and Wang, Wei and Wang, Hao and Yao, Yuncong and Yang, Wankou and Suzuki, Satoshi},
  journal={Journal of Systems Architecture},
  volume={130},
  pages={102675},
  year={2022},
  publisher={Elsevier}
}

@article{reedha2022transformer,
  title={Transformer neural network for weed and crop classification of high resolution UAV images},
  author={Reedha, Reenul and Dericquebourg, Eric and Canals, Raphael and Hafiane, Adel},
  journal={Remote Sensing},
  volume={14},
  number={3},
  pages={592},
  year={2022},
  publisher={MDPI}
}

@article{hendria2023combining,
  title={Combining transformer and CNN for object detection in UAV imagery},
  author={Hendria, Willy Fitra and Phan, Quang Thinh and Adzaka, Fikriansyah and Jeong, Cheol},
  journal={ICT Express},
  volume={9},
  number={2},
  pages={258--263},
  year={2023},
  publisher={Elsevier}
}

@inproceedings{liu2021swin,
  title={Swin transformer: Hierarchical vision transformer using shifted windows},
  author={Liu, Ze and Lin, Yutong and Cao, Yue and Hu, Han and Wei, Yixuan and Zhang, Zheng and Lin, Stephen and Guo, Baining},
  booktitle={Proceedings of the IEEE/CVF international conference on computer vision},
  pages={10012--10022},
  year={2021}
}

@inproceedings{he2016resnet,
  title={Deep residual learning for image recognition},
  author={He, Kaiming and Zhang, Xiangyu and Ren, Shaoqing and Sun, Jian},
  booktitle={Proceedings of the IEEE conference on computer vision and pattern recognition},
  pages={770--778},
  year={2016}
}

@misc{tao2022evaluating,
      title={Evaluating Vision Transformer Methods for Deep Reinforcement Learning from Pixels}, 
      author={Tianxin Tao and Daniele Reda and Michiel van de Panne},
      year={2022},
      eprint={2204.04905},
      archivePrefix={arXiv},
      primaryClass={cs.LG}
}

@InProceedings{pmlr-v87-mueller18a,
  title = 	 {Driving Policy Transfer via Modularity and Abstraction},
  author =       {Mueller, Matthias and Dosovitskiy, Alexey and Ghanem, Bernard and Koltun, Vladlen},
  booktitle = 	 {Proceedings of The 2nd Conference on Robot Learning},
  pages = 	 {1--15},
  year = 	 {2018},
  editor = 	 {Billard, Aude and Dragan, Anca and Peters, Jan and Morimoto, Jun},
  volume = 	 {87},
  series = 	 {Proceedings of Machine Learning Research},
  publisher =    {PMLR},
  url = 	 {https://proceedings.mlr.press/v87/mueller18a.html}
}

@article{Loquercio_2021,
   title={Learning high-speed flight in the wild},
   volume={6},
   ISSN={2470-9476},
   number={59},
   journal={Science Robotics},
   publisher={American Association for the Advancement of Science (AAAS)},
   author={Loquercio, Antonio and Kaufmann, Elia and Ranftl, René and Müller, Matthias and Koltun, Vladlen and Scaramuzza, Davide},
   year={2021},
   month=oct }

@article{loquercio2018dronet,
  title={Dronet: Learning to fly by driving},
  author={Loquercio, Antonio and Maqueda, Ana I and Del-Blanco, Carlos R and Scaramuzza, Davide},
  journal={IEEE Robotics and Automation Letters},
  volume={3},
  number={2},
  pages={1088--1095},
  year={2018},
  publisher={IEEE}
}

@article{dai2020automatic,
  title={Automatic obstacle avoidance of quadrotor UAV via CNN-based learning},
  author={Dai, Xi and Mao, Yuxin and Huang, Tianpeng and Qin, Na and Huang, Deqing and Li, Yanan},
  journal={Neurocomputing},
  volume={402},
  pages={346--358},
  year={2020},
  publisher={Elsevier}
}

@article{
li2023a,
title={A Survey on Transformers in Reinforcement Learning},
author={Wenzhe Li and Hao Luo and Zichuan Lin and Chongjie Zhang and Zongqing Lu and Deheng Ye},
journal={Transactions on Machine Learning Research},
issn={2835-8856},
year={2023},
note={Survey Certification}
}

@article{zhou2019robust,
  title={Robust and efficient quadrotor trajectory generation for fast autonomous flight},
  author={Zhou, Boyu and Gao, Fei and Wang, Luqi and Liu, Chuhao and Shen, Shaojie},
  journal={IEEE Robotics and Automation Letters},
  volume={4},
  number={4},
  pages={3529--3536},
  year={2019},
  publisher={IEEE}
}

@inproceedings{florence2020integrated,
  title={Integrated perception and control at high speed: Evaluating collision avoidance maneuvers without maps},
  author={Florence, Pete and Carter, John and Tedrake, Russ},
  booktitle={Algorithmic Foundations of Robotics XII: Proceedings of the Twelfth Workshop on the Algorithmic Foundations of Robotics},
  pages={304--319},
  year={2020},
  organization={Springer}
}

@article{hansen2021stabilizing,
  title={Stabilizing deep q-learning with convnets and vision transformers under data augmentation},
  author={Hansen, Nicklas and Su, Hao and Wang, Xiaolong},
  journal={Advances in neural information processing systems},
  volume={34},
  pages={3680--3693},
  year={2021}
}

@article{kalantari2022improving,
  title={Improving sample efficiency of value based models using attention and vision transformers},
  author={Kalantari, Amir Ardalan and Amini, Mohammad and Chandar, Sarath and Precup, Doina},
  journal={arXiv preprint arXiv:2202.00710},
  year={2022}
}

@inproceedings{seo2023masked,
  title={Masked world models for visual control},
  author={Seo, Younggyo and Hafner, Danijar and Liu, Hao and Liu, Fangchen and James, Stephen and Lee, Kimin and Abbeel, Pieter},
  booktitle={Conference on Robot Learning},
  pages={1332--1344},
  year={2023},
  organization={PMLR}
}

@inproceedings{tao2023seer,
  title={Seer: Safe efficient exploration for aerial robots using learning to predict information gain},
  author={Tao, Yuezhan and Wu, Yuwei and Li, Beiming and Cladera, Fernando and Zhou, Alex and Thakur, Dinesh and Kumar, Vijay},
  booktitle={2023 IEEE International Conference on Robotics and Automation (ICRA)},
  pages={1235--1241},
  year={2023},
  organization={IEEE}
}

@misc{githubGitHubKumarRoboticskr_mav_control,
	author = {},
	title = {{G}it{H}ub - {K}umar{R}obotics/kr\_mav\_control: {C}ode for quadrotor control --- github.com},
	howpublished = {\url{https://github.com/KumarRobotics/kr_mav_control}},
	year = {},
	note = {[Accessed 14-03-2024]},
}

@inproceedings{meier2015px4,
  title={PX4: A node-based multithreaded open source robotics framework for deeply embedded platforms},
  author={Meier, Lorenz and Honegger, Dominik and Pollefeys, Marc},
  booktitle={2015 IEEE international conference on robotics and automation (ICRA)},
  pages={6235--6240},
  year={2015},
  organization={IEEE}
}

@inproceedings{yu2023avoidbench,
  title={AvoidBench: A high-fidelity vision-based obstacle avoidance benchmarking suite for multi-rotors},
  author={Yu, Hang and de Croon, Guido CH E and De Wagter, Christophe},
  booktitle={2023 IEEE International Conference on Robotics and Automation (ICRA)},
  pages={9183--9189},
  year={2023},
  organization={IEEE}
}

@inproceedings{song2023learning,
  title={Learning perception-aware agile flight in cluttered environments},
  author={Song, Yunlong and Shi, Kexin and Penicka, Robert and Scaramuzza, Davide},
  booktitle={2023 IEEE International Conference on Robotics and Automation (ICRA)},
  pages={1989--1995},
  year={2023},
  organization={IEEE}
}

@article{falanga2019fast,
  title={How fast is too fast? the role of perception latency in high-speed sense and avoid},
  author={Falanga, Davide and Kim, Suseong and Scaramuzza, Davide},
  journal={IEEE Robotics and Automation Letters},
  volume={4},
  number={2},
  pages={1884--1891},
  year={2019},
  publisher={IEEE}
}

@article{ou2021autonomous,
  title={Autonomous quadrotor obstacle avoidance based on dueling double deep recurrent Q-learning with monocular vision},
  author={Ou, Jiajun and Guo, Xiao and Zhu, Ming and Lou, Wenjie},
  journal={Neurocomputing},
  volume={441},
  pages={300--310},
  year={2021},
  publisher={Elsevier}
}

@article{fastplanner,
  title={Robust and efficient quadrotor trajectory generation for fast autonomous flight},
  author={Zhou, Boyu and Gao, Fei and Wang, Luqi and Liu, Chuhao and Shen, Shaojie},
  journal={IEEE Robotics and Automation Letters},
  volume={4},
  number={4},
  pages={3529--3536},
  year={2019},
  publisher={IEEE}
}

@misc{yuwei-wu-fastplanner-github,  
  author       = {Yuwei Wu},  
  title        = {Fast-Planner},  
  year         = {2023},  
  url          = {https://github.com/yuwei-wu/Fast-Planner},  
  note         = {Accessed: 2024-07-01}  
}

@INPROCEEDINGS{doubledescription,
  author={Wang, Zhepei and Ye, Hongkai and Xu, Chao and Gao, Fei},
  booktitle={2021 IEEE International Conference on Robotics and Automation (ICRA)},
  title={Generating Large-Scale Trajectories Efficiently using Double Descriptions of Polynomials},
  year={2021},
  volume={},
  number={},
  pages={7436-7442},
  doi={10.1109/ICRA48506.2021.9561585}}

@misc{yuwei-wu-doubledescription-github,  
  author       = {Yuwei Wu},  
  title        = {Double Description},  
  year         = {2023},  
  url          = {https://github.com/yuwei-wu/DoubleDescription/tree/dodge},  
  note         = {Accessed: 2024-07-01}  
}

@article{bhattacharya2024vision,
  title={Vision Transformers for End-to-End Vision-Based Quadrotor Obstacle Avoidance},
  author={Bhattacharya, Anish and Rao, Nishanth and Parikh, Dhruv and Kunapuli, Pratik and Matni, Nikolai and Kumar, Vijay},
  journal={arXiv preprint arXiv:2405.10391},
  year={2024}
}

@inproceedings{bhattacharya2024monocular,
  title={Monocular Event-Based Vision for Obstacle Avoidance with a Quadrotor},
  author={Bhattacharya, Anish and Cannici, Marco and Rao, Nishanth and Tao, Yuezhan and Kumar, Vijay and Matni, Nikolai and Scaramuzza, Davide},
  booktitle={8th Annual Conference on Robot Learning},
  year={2024}
}
